\documentclass[11pt]{article}

\usepackage[final]{acl}

\usepackage{times}
\usepackage{latexsym}

\usepackage{amsfonts}
\usepackage[T1]{fontenc}

\usepackage[utf8]{inputenc}
\usepackage{multirow}
\usepackage[most]{tcolorbox}
\usepackage{microtype}
\usepackage{subcaption}
\usepackage{amsmath}
\usepackage{booktabs}

\definecolor{rowblue}{RGB}{220,230,240}
\usepackage[table]{xcolor}
\definecolor{casebluebg}{rgb}{0.8,0.9,1}
\newcommand{\best}[1]{\textbf{#1}}
\newcommand{\second}[1]{\underline{#1}}

\usepackage{makecell}
\usepackage{pifont}
\newcommand{\cmark}{\textcolor{green!60!black}{\ding{51}}}
\newcommand{\xmark}{\textcolor{red}{\ding{55}}}

\usepackage{xcolor}

\definecolor{v1bg}{RGB}{255,248,230}
\definecolor{v1border}{RGB}{210,140,30}

\definecolor{v2bg}{RGB}{255,235,235}
\definecolor{v2border}{RGB}{190,40,40}

\definecolor{v3bg}{RGB}{230,248,245}
\definecolor{v3border}{RGB}{25,150,130}

\definecolor{v4bg}{RGB}{245,235,255}
\definecolor{v4border}{RGB}{120,60,200}

\definecolor{v5bg}{RGB}{230,240,255}
\definecolor{v5border}{RGB}{30,90,180}
\usepackage{inconsolata}

\usepackage{graphicx}
\usepackage{array}

\newcommand{\name}{\emph{OmniLayout}}
\newcommand{\tai}{\textcolor{black}}
\definecolor{casegreenbg}{RGB}{240,248,240}      
\definecolor{casegreenborder}{RGB}{34,139,34}    
\title{OmniLayout: A Schematic-Coupled Multimodal Benchmark for Constraint-Aware Geometric Reasoning in PCB Layout}

\author{
 \textbf{Taiting Lu\textsuperscript{1}},
 \textbf{Kaiyuan Lin\textsuperscript{1}},
 \textbf{Mingjia Wang\textsuperscript{2}},
  \textbf{Haolin Ye\textsuperscript{1}},
  \textbf{Runze Liu\textsuperscript{1}},
   \textbf{Yuxin Tian\textsuperscript{3}},
\\
 \textbf{Vahe Melkonyan\textsuperscript{2}},
 \textbf{Haoyu Wang\textsuperscript{2}},
 \textbf{Muchuan Wang\textsuperscript{1}},
 \textbf{Chufan Hong \textsuperscript{3}},
  \textbf{Yifan Yang \textsuperscript{4}},
\\
  \textbf{Sung-Liang Chen \textsuperscript{2}},
\textbf{Yi-Chao Chen \textsuperscript{2}},
\textbf{Yincheng Jin\textsuperscript{5}},
\textbf{Mahanth Gowda\textsuperscript{1\textdagger{}}},
\\
 \textsuperscript{1}Pennsylvania State University,
 \textsuperscript{2}Shanghai Jiao Tong University,
 \textsuperscript{3}Independent Researcher,
\\
 \textsuperscript{4}Microsoft Research,
 \textsuperscript{5}Binghamton University
}

\definecolor{caseblueborder}{RGB}{0,112,192}
\begin{document}

\maketitle

\begingroup
\renewcommand\thefootnote{}
\footnotetext{\textdagger{} Corresponding author.}
\footnotetext{Project Page: \url{https://www.omnieda.com/}}
\endgroup

\begin{abstract}
Recent large multimodal models (LMMs) have demonstrated remarkable progress in 3D spatial reasoning, spatial grounding, and fine-grained geometric understanding. However, their ability to reason about densely packed object placement under strict spatial and functional constraints remains largely unexplored, despite being a fundamental challenge in practical electronic design automation (EDA) workflows. To bridge this gap, we introduce \textbf{\name}, the first benchmark designed to evaluate LLMs on printed-circuit-board (PCB) layout placement reasoning under real-world geometric, routing, and connectivity constraints.
{\name} contains \tai{1,681} industrial-grade schematic-coupled PCB layouts and includes four tasks: \textbf{(1) geometric reasoning} for IC physical placement, with \tai{77.24\textit{K}} placement instances constrained within PCB board boundaries; \textbf{(2) routability-aware placement reasoning}, generating physically valid component placements;
\textbf{(3) relational preservation and reference agreement}, measuring
whether placements preserve schematic-derived spatial relationships and
reporting agreement with an expert reference layout as an auxiliary
diagnostic; and \textbf{(4) tool-augmented agentic reasoning}in which models invoke
external tools to accomplish Tasks (1)--(3).
Our results reveal substantial limitations of current LMMs in PCB layout placement, including weak geometric reasoning, poor routability optimization, and only partial preservation of schematic-derived relationships.

\end{abstract}

\section{Introduction}\label{intro}

\begin{figure}[h]
  \centering
  \includegraphics[width=\columnwidth]{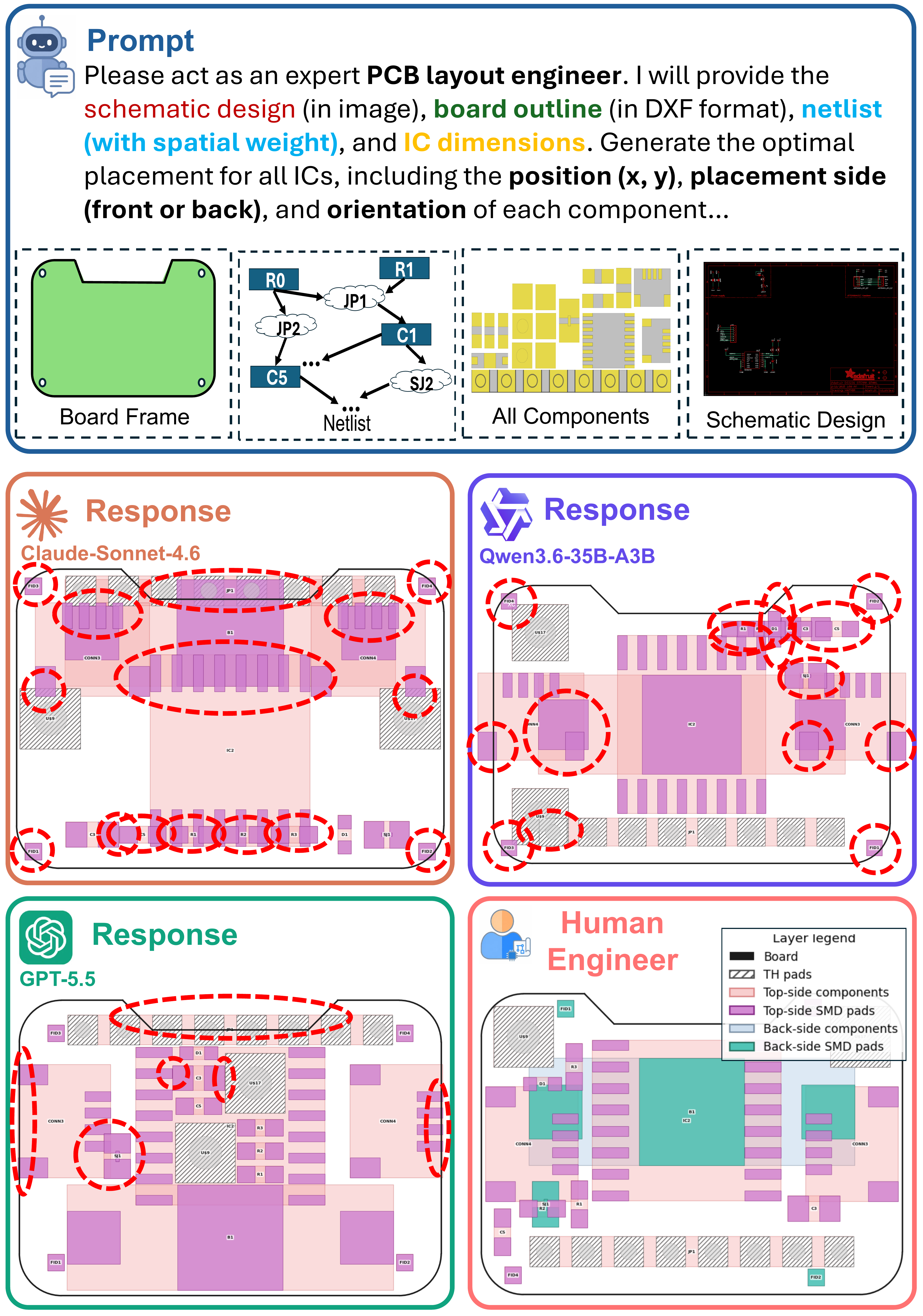}
  \caption{\textbf{Large multimodal models (LMMs) fail to perform reliable PCB layout placement under geometric and electrical constraints.} Frequent overlap violations, highlighted by dashed \textcolor{red}{red} circles,
reveal substantial weaknesses in LMM spatial reasoning and
constraint-aware layout generation.}
  
  \label{fig:lmm_test}
  \vspace{-0.25in}
\end{figure}

Driven by the explosive growth in semiconductor technology, modern Printed Circuit Boards (PCBs) now contain hundreds to thousands of components that must be precisely positioned and electrically connected within a limited area.
Despite this growing complexity, the placement and routing (PnR) process, such as the example shown in Figure \ref{fig:lmm_test}, remains largely manual, with experienced layout engineers spending several weeks to complete a single design. 
However, real-world PCB layout designs are almost exclusively proprietary and rarely publicly available. Even within the academic community, most PCB PnR works \cite{zhang2025cypress,chen2025clearance,han2026pcb,han2026moduplace} are trained and evaluated on closed-source private datasets containing only 10 to 20 samples, severely hindering reproducible benchmarking and generalization.

\begin{table}[h!]
\centering
\caption{
Comparison of publicly available PCB layout benchmarks and datasets.
}
\label{tab:pcb_benchmark}
\begingroup
\fontsize{5.5}{6.5}\selectfont
\setlength{\tabcolsep}{3pt}
\renewcommand{\arraystretch}{1.15}
\begin{tabular}{
m{1.1cm}
>{\centering\arraybackslash}m{0.4cm}
>{\centering\arraybackslash}m{0.4cm}
>{\centering\arraybackslash}m{0.3cm}
>{\centering\arraybackslash}m{0.5cm}
>{\centering\arraybackslash}m{0.45cm}
>{\centering\arraybackslash}m{0.48cm}
>{\centering\arraybackslash}m{0.4cm}
>{\centering\arraybackslash}m{0.4cm}
>{\centering\arraybackslash}m{0.45cm}
>{\centering\arraybackslash}m{0.4cm}
}
\toprule
\multirow{2}{*}{Dataset} &
\multirow{2}{*}{Layout} &
\multirow{2}{*}{\makecell{Ref.\\Layout}} &
\multirow{2}{*}{Sch.} &
\multicolumn{3}{c}{Annotation} &
\multirow{2}{*}{\makecell{Sem.\\Attr.}} &
\multirow{2}{*}{DRV} &
\multirow{2}{*}{\#Layers} &
\multirow{2}{*}{Year} \\
\cmidrule(lr){5-7}
& & & & \#Comp. & \#Pins & \#Nets & & & & \\
\midrule
Cypress~\cite{zhang2025cypress}
& 10 & \xmark & \xmark & 1.2K & 9.5K & 2.6K & \xmark & \xmark & 2 & 2025 \\
\rowcolor{rowblue}
\textbf{{\name} (Ours)}
& \textbf{1681} & \cmark & \cmark & \textbf{77.24K}
& \textbf{168.8K} & \textbf{139.4K} & \cmark & \cmark & 2-8 & 2026 \\
\bottomrule
\end{tabular}
\endgroup
\end{table}

In recent years, large multimodal models (LMMs)  \cite{mouselinos2024beyond,chen2024spatialvlm,cheng2024spatialrgpt,rodionov2025floorplanqa,huang2025fireplace,lu2026omniroutingsemanticcoupledmultimodalbenchmark}, have demonstrated strong capabilities in 3D spatial reasoning, fine-grained geometry, spatial grounding, and spatial planning. 
These advances suggest new opportunities for applying LMMs to engineering workflows, particularly for PCB PnR under geometric and electrical constraints. 
However, these existing spatial-LMM works largely overlook the fundamental challenge of placing densely packed objects within tightly constrained spaces while simultaneously satisfying complex geometric and functional requirements. This raises a critical open question: \textbf{\textit{Can current LMMs generate PCB layouts that are geometrically legal,
routing-aware, and consistent with schematic-derived circuit
relationships across diverse real-world designs?}}

To answer this question, we designed a set of experiments to evaluate the PCB PnR capabilities of several state-of-the-art LMMs, including Claude-Sonnet-4.6 \cite{anthropic_claude_opus_47}, GPT-5.5 \cite{openai_gpt55}, and Qwen3.6-35B-A3B \cite{qwen3-35b-a3b}, benchmarked against experienced human layout engineers. Specifically, we evaluated their ability to perform component placement under the geometric and electrical constraints characteristic of real-world PCB designs. As illustrated in Figure \ref{fig:lmm_test}, all evaluated models exhibit more than 5 component overlap errors on PCB layout tasks, while consistently confining all components to a single board side rather than leveraging the spatial capacity of the board.
These results reveal that current LMMs fundamentally lack the spatial constraint reasoning and dense placement optimization capabilities required for PCB PnR, as they fail to satisfy component non-overlap constraints, multi-layer spatial utilization, and electrical connectivity requirements essential to practical layout design.

\begin{figure*}[h]
    \centering
    \includegraphics[width=0.9\textwidth]{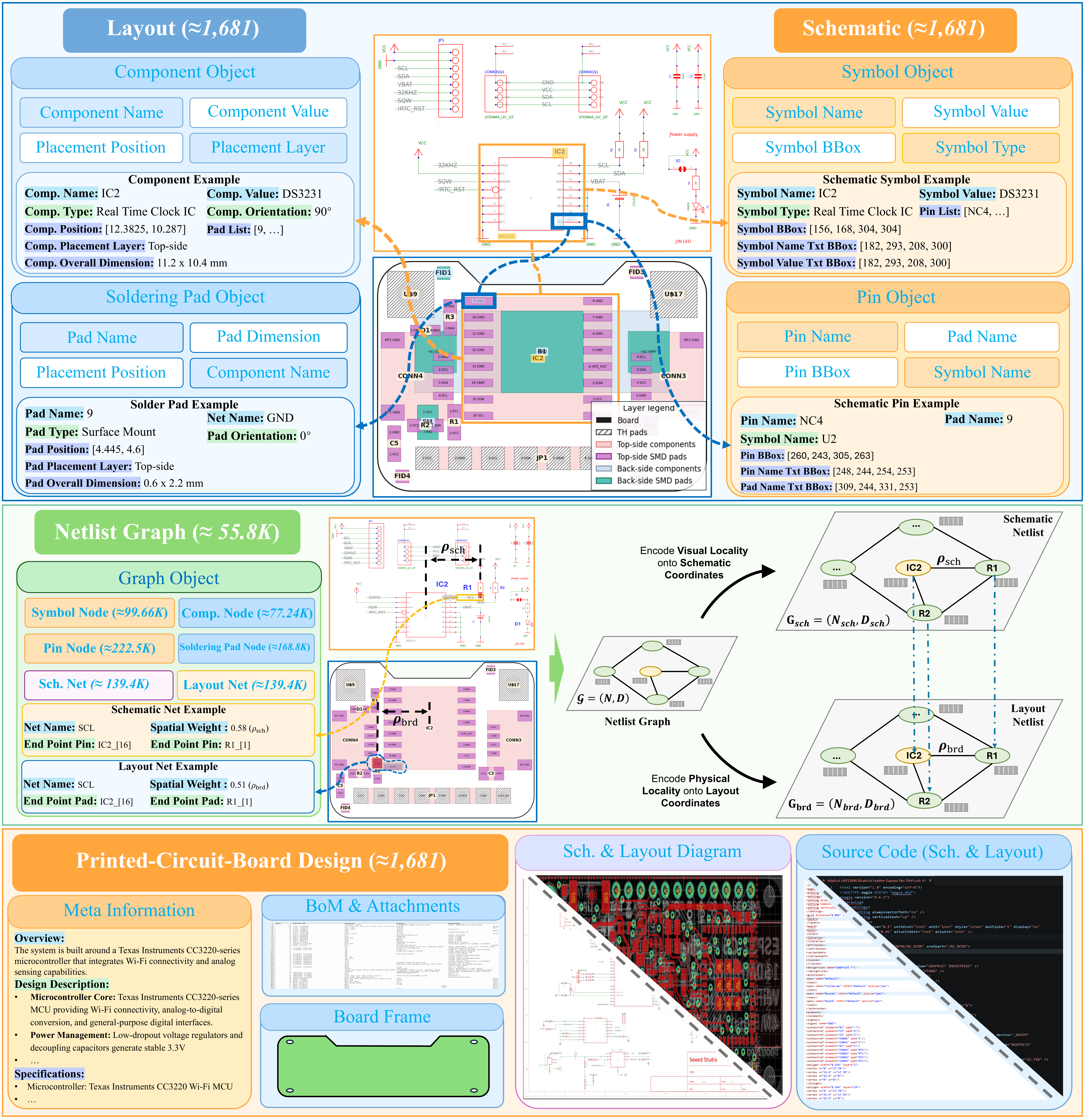}
    \caption{Overview of {\name} benchmark with representative cases.}
    \label{fig:dataset_overview}
    \vspace{-0.2in}
\end{figure*}

To bridge this gap, we introduce \textbf{\name}, a large-scale benchmark of real-world schematic-coupled PCB layout designs for evaluating LMMs on multimodal constraint-aware geometric reasoning for PCB placement tasks. 
As shown in Figure \ref{fig:dataset_overview}, {\name} contains \tai{1,681} real-world PCB layouts paired with corresponding schematics and expert-designed reference layouts, covering diverse board outlines ranging from industrial PCB stackups from 2 to 8 layers and board areas range from 31.53\,mm$^2$ to 22,458.93\,mm$^2$. 
The benchmark includes \tai{77.24\textit{K}} IC and component placement instances, \tai{168.82\textit{K}} SMDs and pads, and \tai{55.83\textit{K}} netlist graphs with spatial distance weights and connectivity semantic attributes.
We first propose a systematic evaluation protocol for assessing PCB layout designs generated by LMMs with respect to routability, geometric legality, and electrical functionality.
Our benchmark evaluates four core abilities, including 
\textbf{(i) geometric reasoning} for physically valid IC and component placement within PCB board boundaries, 
\textbf{(ii) routability-aware placement reasoning} for generating routing-efficient layouts under netlist constraints, 
\textbf{(iii) relational-preservation reasoning} for retaining
schematic-derived spatial relationships and functional grouping, while
using similarity to an expert layout as an auxiliary reference
anchor,
and \textbf{(iv) tool-augmented agentic reasoning} for iterative PCB placement optimization using external layout-status visualization, overlap localization, geometric-constraint checking and HPWL-based routing feedback, and routability analysis tools to guide constraint-aware layout refinement.

Our main contributions are as follows: 
\textbf{(i)}: We introduce \textbf{\name}, a comprehensive benchmark for evaluating LMMs on PCB layout placement, containing \tai{1,681} real-world schematic-coupled PCB layouts with expert-designed reference placements and systematic evaluation protocols.
\textbf{(ii)} We systematically evaluate state-of-the-art LMMs on PCB layout placement under multiple settings, including single-zeroshot placement generation and tool-augmented agentic refinement with iterative visual, geometric, and routability feedback.
\textbf{(iii)} We provide a detailed analysis of LMM capabilities in PCB layout placement, examining performance across geometric legality, routability-aware spatial reasoning, preservation of schematic-derived relationships and agreement with an
expert reference layout, and the effectiveness of tool-augmented agentic refinement with iterative layout feedback.

\section{Related Work}\label{related_work}

\begin{figure*}[h]
    \centering
    \begin{minipage}{0.249\textwidth}
        \centering
        \includegraphics[width=0.8\linewidth]{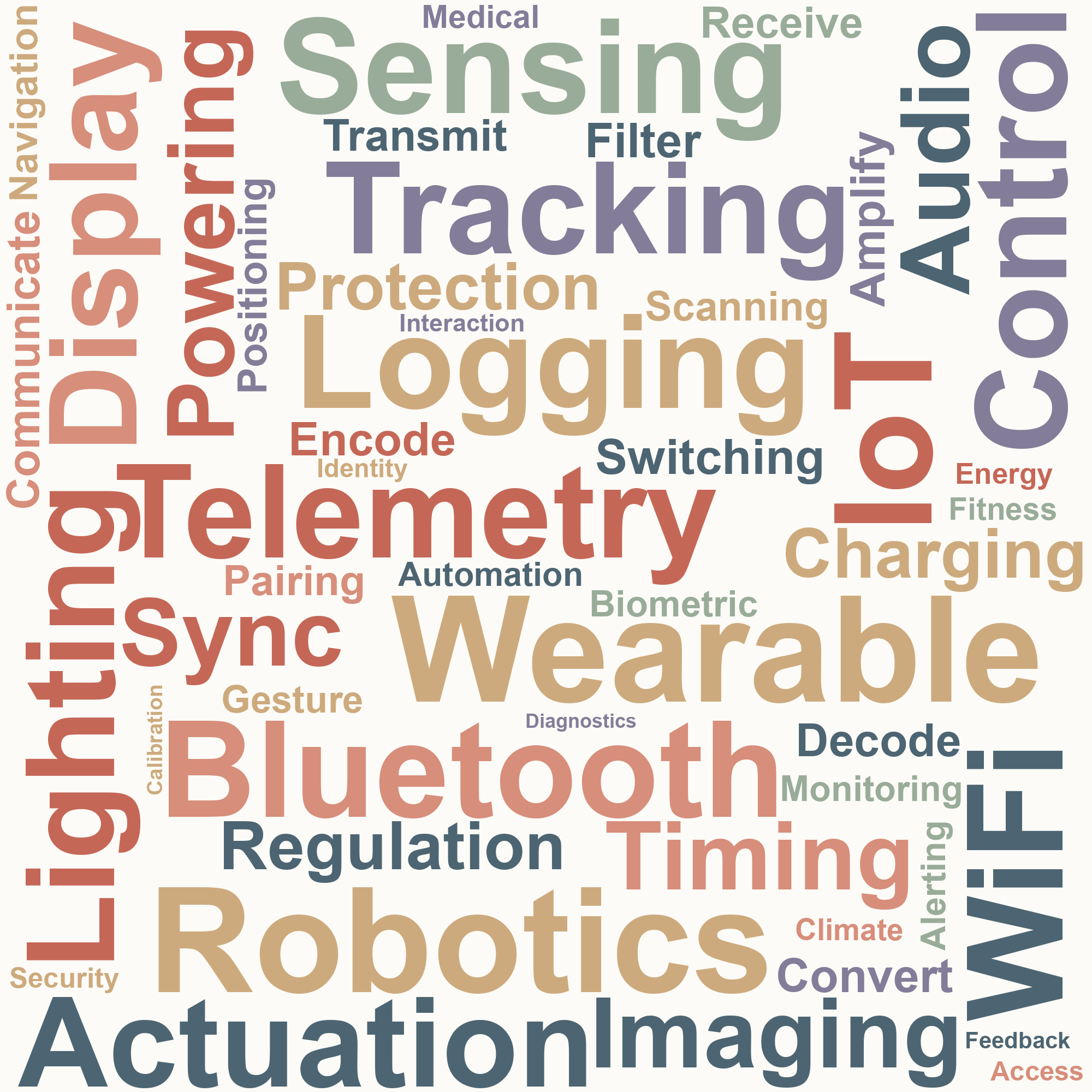}
        \subcaption{wordcloud of topics}
    \end{minipage}%
    \begin{minipage}{0.249\textwidth}
        \centering
        \includegraphics[width=\linewidth]{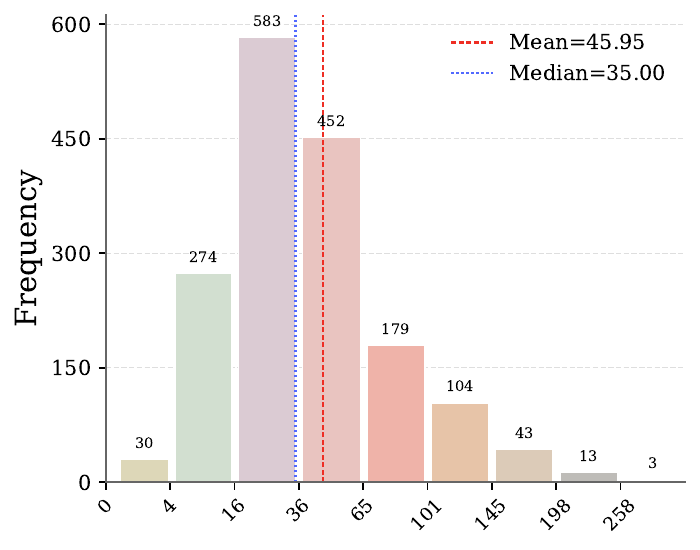}
        \subcaption{Components per diagram.}
    \end{minipage}%
    \begin{minipage}{0.249\textwidth}
        \centering
        \includegraphics[width=\linewidth]{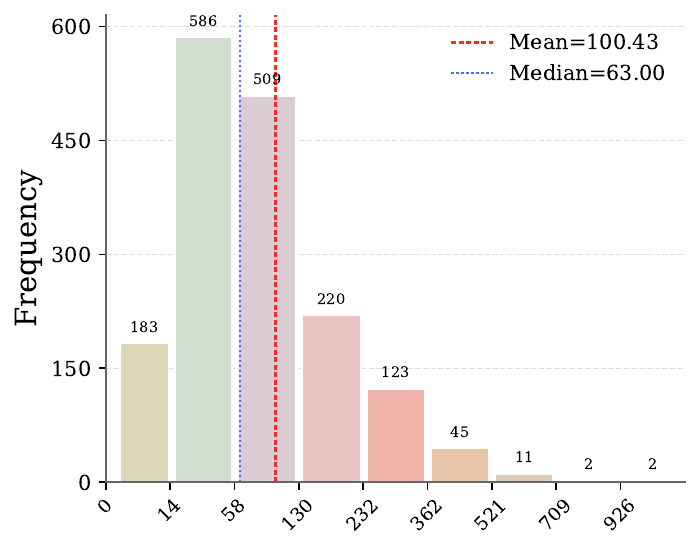}
        \subcaption{Pins per diagram.}
    \end{minipage}%
    \begin{minipage}{0.249\textwidth}
        \centering
        \includegraphics[width=\linewidth]{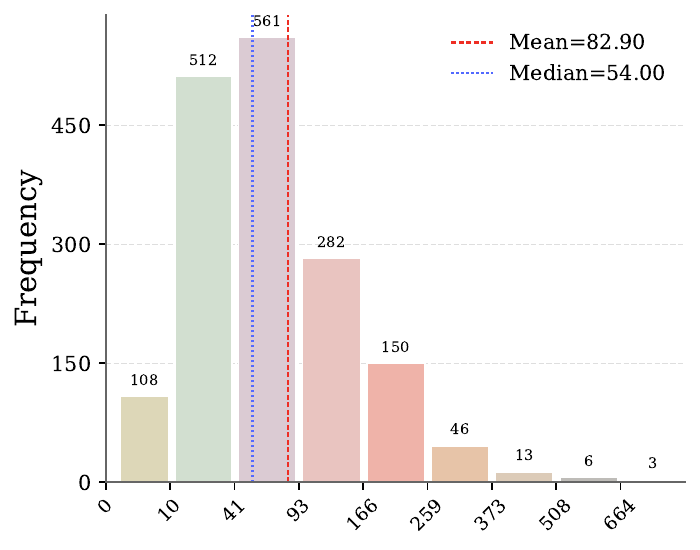}
        \subcaption{Nets per diagram.}
    \end{minipage}
     \vspace{-0.1in}
    \caption{Statistical overview of the {\name} benchmark. The dataset covers a wide range of electronic domains, with each layout design containing up to \tai{327 components, 1,172 pins, and 840 nets.}}
    \vspace{-0.2in}
    \label{fig:img2sch_statistic}
\end{figure*}

\textbf{Spatial Reasoning in LMMs.}
LMMs such as SpatialVLM, SpatialRGPT, FloorplanQA, and Fireplace~\cite{chen2024spatialvlm,cheng2024spatialrgpt,rodionov2025floorplanqa,huang2025fireplace,li2025industrynav} have demonstrated increasingly strong spatial reasoning capabilities, progressing from symbolic and textual spatial reasoning benchmarks such as SPARTQA, SpaRTUN, and StepGame \cite{mirzaee2021spartqa,mirzaee2022transfer,shi2022stepgame} toward multimodal and grounded spatial understanding in benchmarks such as SpatialEval, VSI-Bench, Open3DVQA, 3DSRBench, and CA-VQA \cite{yang2025thinking,zhang2025open3d,ma20253dsrbench}. Recent work further explores spatial world modeling through cognitive-map prompting, depth-aware multimodal learning, and neuro-symbolic reasoning \cite{yang2025thinking,cai2025spatialbot,ali2025neuro}. Despite these advances, existing LMM research has largely overlooked the challenge of placing densely packed objects within tight spatial boundaries while simultaneously satisfying complex geometric and functional constraints.

\textbf{Benchmarks for PCB Placement.} 
In recent years, a growing number of both commercial tools \cite{kuo2006pcb,jones2023distributed} and academic methods \cite{zhang2025cypress,chen2025clearance,han2026pcb,han2026moduplace} have been proposed to automate the PCB component placement process, spanning classical optimization algorithms, constraint-based solvers, and emerging learning-based approaches.
Representative commercial tools such as Cadence Allegro X AI \cite{kuo2006pcb,jones2023distributed}, DeepPcb \cite{deeppcb2026} and Quilter \cite{quilter2026} adopt search-based and reinforcement learning strategies respectively to automate PCB component placement. 
However, these approaches are developed on proprietary private datasets that are never publicly released, and generating a single placement candidate can take up to several hours, posing significant barriers to both reproducibility and practical deployment.
Cypress \cite{zhang2025cypress} and ModuPlace \cite{han2026moduplace} represent the most recent academic efforts, proposing a VLSI-inspired GPU-accelerated placement framework and an LLM-assisted modular placement method respectively, yet both are evaluated on small closed-source or synthesized benchmarks of fewer than 100 components.
Current works remain severely limited in both scale and accessibility, as most methods are evaluated on closed-source private datasets containing only tens of designs with limited numbers of components, far below the complexity of real-world industrial PCBs, which fundamentally hinders the generalizability and reproducibility of proposed methods across diverse circuit designs.

\section{Benchmark Construction}\label{benchmark}

\subsection{Task Formulation}
To provide a comprehensive evaluation framework for PCB placement tasks, our task formulation covers four key capabilities of constraint-aware geometric reasoning:
\textbf{(i)} \textbf{Spatial Reasoning}, which requires determining the position, orientation, and layer assignment of a large number of densely packed components within an irregularly shaped board boundary while satisfying geometric spacing and non-overlap constraints;
\textbf{(ii)} \textbf{Routability}, which evaluates whether the resulting component placement can be successfully connected by electrical traces without conflicts, a necessary condition for a functional PCB;
\textbf{(iii) Relational Preservation and Reference Agreement}, which
evaluates whether a predicted placement preserves spatial relationships
derived from the schematic and reports its similarity to an expert
reference layout. Because PCB placement admits multiple valid solutions,
reference-layout similarity is treated as an auxiliary diagnostic;
\textbf{(iv) Agentic Tool Use for Constraint-Aware Geometric Reasoning},
which formulates PCB component placement as an iterative decision-making
problem in which LMMs use visualization, overlap checking, boundary
checking, and routing feedback to refine their placements.

\subsection{Source-driven Annotation Curation}
We construct the {\name} benchmark through a scalable automated pipeline, with full curation details in \tai{Appendix \ref{sec:appendix:data}}. 
We gather real-world schematic designs from open-source hardware platforms such as SparkFun, Arduino, Adafruit, and GitHub~\cite{sparkfun, github, arduino, adafruit, seeedstudio, protocentral}. The collected projects are implemented in Autodesk EAGLE~\cite{autodesk_eagle}, a widely used industrial EDA tool that adopts an XML-based file format, allowing efficient parsing of circuit structures and connectivity information through programmatic extraction.
To evaluate PCB placement across spatial, routability, and electrical functionality dimensions, we require two types of ground-truth annotations: \tai{(i)} schematic visual positions encoding intended electrical relationships between components, and \tai{(ii)} reference layout annotations capturing ground-truth component positions, orientations, layer assignments, and board frame geometry. 
Rather than costly manual annotation, we leverage two complementary EDA rendering engines.
For schematics, we adopt the rendering engine from \cite{lu2026omnisch}, which parses EAGLE XML files to produce high-fidelity schematic diagrams with pixel-aligned component positions and net-level connectivity annotations, serving as reference signals for evaluating electrical functionality preservation. 
For reference layouts, while component placements can be extracted directly from EAGLE software, board frame geometry is not captured; we therefore develop a dedicated PCB layout rendering engine that parses source-driven annotations and board frame geometry directly from EAGLE XML layout files containing the reference layout information.
Beyond annotation extraction, this engine additionally serves as an interactive visualization tool within our agentic evaluation framework (Section~\ref{sec:exp}), rendering the current placement state as visual feedback to guide LMMs in iteratively refining component positions.

\subsection{Statistics of OmniLayout Benchmark} 
As illustrated in Figure~\ref{fig:dataset_overview} and Figure~\ref{fig:img2sch_statistic}, {\name} comprises \tai{1,681} real-world schematic-coupled PCB layout designs spanning diverse application domains (e.g., robotics, wireless systems, and sensing), providing a rich and diverse resource for supervised learning in PCB placement and other EDA tasks. 
Our annotations are organized into three levels:
\textbf{(i) entity-level annotations} for \tai{77.24K}
component placements and \tai{168.82K} pads, together with their
corresponding schematic symbols and pins;
\textbf{(ii) netlist-graph annotations}, comprising \tai{55.83K}
graphs that encode circuit topology and connectivity; and
\textbf{(iii) spatially weighted net annotations}, comprising
\tai{139.4K} nets represented in both schematic and expert-layout
coordinate frames.
The \tai{55.83K} figure denotes netlist graphs, whereas \tai{139.4K} denotes the nets contained within those graphs. Annotation fields and representations are provided in Appendix~\ref{sec:annotation_schema}.
These representations support two relational evaluations: schematic
relational distance, which measures agreement with spatial relationships
derived from the schematic, and reference-layout similarity, which
measures agreement with one expert placement.

\begin{figure*}[t]
\vspace{-0.1in}
    \centering
    \includegraphics[width=0.98\textwidth]{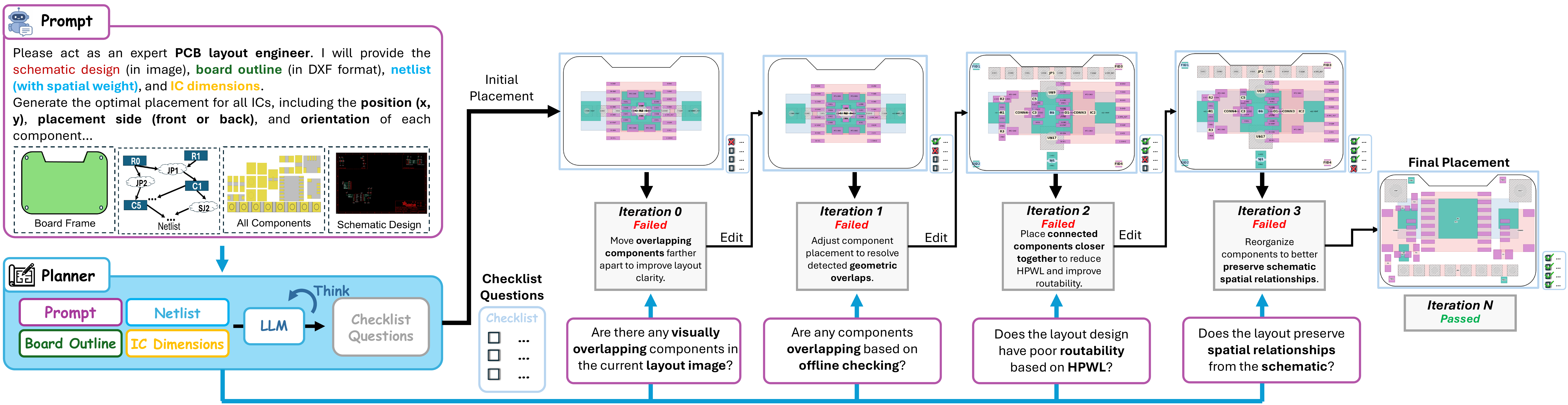}
    \caption{Overview of a multi-modal, multi-agent, multi-round agentic framework for evaluating LMM-based tool usage in PCB component placement.}
    \label{fig:react_framework}
    \vspace{-0.25in}
\end{figure*}

\section{Evaluation Metrics}

\subsection{Evaluation Protocol}
To the best of our knowledge, there is no standardized protocol 
for evaluating PCB component placement predictions from LMMs 
that jointly considers geometric legality, routability, and preservation
of schematic-derived spatial relationships.
PCB PnR task is a well-known non-convex NP-hard problem: 
given $n$ components to be placed within a constrained board area, 
the solution space grows combinatorially due to discrete orientations, layer assignments, and non-overlapping constraints, yielding multiple feasible solutions with no efficient algorithm guaranteed to find the global optimum. 
Consequently, a PCB design generally admits multiple valid placements,
and an expert reference layout should not be treated as a unique
ground-truth solution.
To evaluate a predicted placement against a reference design, 
we formulate PCB placement evaluation as a constrained 
optimization assessment problem. Given a set of components $V$, 
nets $E$, and board boundary $B$, the placement quality is 
measured by:
\vspace{-0.5em}
\begin{equation}
\small
\begin{aligned}
\min_{\mathbf{x},\mathbf{y},\boldsymbol{\theta},\mathbf{l}}
\quad
& W(\mathbf{x},\mathbf{y},\boldsymbol{\theta})
+ \lambda_1 C_{\mathrm{p2p}}(\mathbf{x},\mathbf{y})
\\
& + \lambda_2 D_{\mathrm{geo}}
(\mathbf{x},\mathbf{y},\boldsymbol{\theta},\mathbf{l})
\end{aligned}
\label{eq:placement}
\end{equation}
subject to
{\small
$\theta_i \in \{0^\circ,90^\circ,180^\circ,270^\circ\}$,
$l_i \in \{\text{top},\text{back}\}$, and
$(x_i,y_i)\in B$, $\forall v_i\in V$},
where
{\small
$W(\mathbf{x},\mathbf{y},\boldsymbol{\theta})$}
measures total wirelength as a proxy for routing efficiency,
{\small
$C_{\mathrm{p2p}}(\mathbf{x},\mathbf{y})$}
penalizes pad-to-pad clearance violations, and
{\small
$D_{\mathrm{geo}}
(\mathbf{x},\mathbf{y},\boldsymbol{\theta},\mathbf{l})$}
penalizes component overlaps between components assigned to the same
board side. The coefficients $\lambda_1$ and $\lambda_2$ control the
relative weights of the clearance and overlap penalties with respect
to the wirelength objective. Because PCB placement admits multiple valid solutions,
reference-layout agreement is not included in the primary placement
objective. Instead, it is reported separately as an auxiliary
diagnostic:

\vspace{-0.5em}
\begin{equation}
\small
D_{\mathrm{ref}}
(\mathbf{x},\mathbf{y},\mathbf{x}^{*},\mathbf{y}^{*})
=
\frac{1}{|E|}
\sum_{(v_i,v_j)\in E}
\left\|
\Delta_{ij}^{\mathrm{pred}}
-
\Delta_{ij}^{\mathrm{ref}}
\right\|_2
\label{eq:reference_distance}
\end{equation}
\vspace{-0.5em}

where
{\small
$\Delta_{ij}^{\mathrm{pred}}
=(x_i-x_j,y_i-y_j)$}
and
{\small
$\Delta_{ij}^{\mathrm{ref}}
=(x_i^{*}-x_j^{*},y_i^{*}-y_j^{*})$}
denote the relative spatial displacements of connected component pair
$(v_i,v_j)$ in the predicted placement and the expert reference layout,
respectively. Lower $D_{\mathrm{ref}}$ indicates closer agreement with
the relative spatial patterns found in the expert reference. Because
PCB placement admits multiple valid solutions, this metric is reported
only as an auxiliary diagnostic and should not be interpreted as a
measure of electrical correctness or overall placement quality. For readability, the result tables report the corresponding bounded
reference-layout similarity:
\vspace{-0.5em}
\begin{equation}
\small
S_{\mathrm{ref}}
=
\frac{1}{1+D_{\mathrm{ref}}},
\label{eq:reference_similarity}
\end{equation}
\vspace{-0.5em}
where higher values indicate greater agreement with the expert reference
and the reference layout itself has $S_{\mathrm{ref}}=1$.

\subsection{Evaluation Criteria}

We define evaluation metrics across four complementary dimensions to comprehensively assess PCB placement quality; full details of each metric are provided in \tai{Appendix \ref{sec:appendix:eval}} due to space constraints.
\textbf{(i)} \textbf{Geometric Legality}, which evaluates whether the predicted placement satisfies physical constraints, including normalized component overlap area, out-of-boundary ratio, and runtime, collectively measuring whether the placement is physically valid within board boundary $B$.
\textbf{(ii)} \textbf{Routability}, which assesses routing quality of the predicted placement across both pre-routing proxies \textemdash{} half-perimeter wirelength (HPWL)~\cite{naylor2001non}, net crossing (NC)~\cite{zhang2025cypress}, and net separation~\cite{cheng2022net} \textemdash{} and post-routing metrics obtained via FreeRouting~\cite{freerouting2024}, including actual routed wirelength (PRWL), routability ratio~\cite{zhang2025cypress} and number of vias (\#Vias).
\textbf{(iii)} \textbf{Electrical Functionality Preservation}, which measures whether the predicted placement preserves the relative spatial relationships of subcircuits defined in the circuit schematic. Since PCB placement does not have a unique ground truth solution, we evaluate the degree to which the predicted layout aligns with the subcircuit topology of the schematic diagram in terms of visual relative distances, as well as the component-level relative placement distances of the expert-designed reference layout.
\textbf{(iv)} \textbf{Agentic Placement}, which evaluates LMMs iteratively refining PCB placement through tool use, measured by: (a) \textit{Constraint Satisfaction Rate}, the ratio of geometric and electrical constraints satisfied at the final placement iteration; (b) \textit{Step Efficiency}, the average number of tool-use steps required to reach a valid placement; and (c) \textit{Placement Refinement Quality}, whether successive placement updates progressively reduce overlap violations, out-of-boundary components, and routing conflicts across iterations.

\section{Experiment and Findings}

\subsection{Experimental Setup}\label{sec:exp}

\textbf{Study Setup.} The tested LMMs in this section include GPT-5.5~\cite{openai_gpt55}, GPT-5-mini~\cite{openai_gpt5mini}, Gemini-3.1-Pro-Preview~\cite{google_gemini3_pro}, Gemini-2.5-Flash-Lite~\cite{google_gemini_25_flash}, Claude-Sonnet-4.6~\cite{anthropic_claude_opus_47}, LLaMA-4-Maverick~\cite{meta_llama4}, Mistral-3-14B~\cite{mistral_medium35}, Qwen3.5-9B~\cite{qwen35_9b}, and Qwen3.6-35B-A3B~\cite{qwen3-35b-a3b}. We evaluate these models alongside one commercial and five open-source PCB placement baselines. We conduct two studies: \textbf{(i)} \textbf{Zero-shot Evaluation}, where all LMMs are prompted once without intermediate interaction to assess end-to-end PCB component placement. All settings include the board boundary, netlist, and component dimensions, while we further study the impact of additional information, including schematic images, schematic-aware netlist graphs, global schematic positions, and few-shot examples; and \textbf{(ii)} \textbf{Agentic Evaluation}, where LMMs are equipped with layout visualization, overlap checking, and routability verification tools to iteratively refine component placement under geometric and electrical constraints. Implementation details are provided in \tai{Appendix \ref{sec:appendix:exp}}.

\begin{table*}[!ht]
\centering
\caption{\textbf{Oneshot evaluation of LMMs on PCB component placement.} \textit{OO} denotes total overlapping area between components; \textit{OoB} denotes components placed outside the board boundary; \textit{RT} denotes inference time in seconds; \textit{HPWL} denotes half-perimeter wirelength; \textit{NC} denotes number of crossings; \textit{NS} denotes average net separation distance; \textit{PRWL} denotes predicted routed wirelength; \textit{RR} denotes fraction of successfully routed nets; \textit{\#Vias} denotes number of vias; \textit{Sch. SW-Net} denotes spatial-weight netlist extracted from schematic; \textit{Sch. Img} denotes schematic image; \textit{Sch. Sem.} denotes schematic semantic attributes for components; \textit{Sch.} denotes schematic relational distance;
\textit{Ref.} denotes reference-layout similarity.}
\vspace{-0.05in}
\label{tab:zeroshot}

\begingroup
\tiny
\setlength{\tabcolsep}{4pt}
\renewcommand{\arraystretch}{0.40}

\resizebox{\textwidth}{!}{%
\begin{tabular}{l c c c c c c c c c c c c c c c c}
\toprule

\noalign{\vspace{-2pt}}
\multirow{2}{*}{Model}
& \multirow{2}{*}{Size}
& \multirow{2}{*}{\shortstack{Input\\Data}}
& \multicolumn{4}{c}{Geometric Legality}
& \multicolumn{7}{c}{Routability}
&\multicolumn{2}{c}{Relational Agreement}
& \multicolumn{1}{c}{Overall} \\

\cmidrule(lr){4-7}
\cmidrule(lr){8-14}
\cmidrule(lr){15-17}

& &
& \shortstack{OO$\downarrow$}
& \shortstack{OoB$\downarrow$}
& \shortstack{RT$\downarrow$}
& \shortstack{P-Pass(\%)$\uparrow$}
& \shortstack{HPWL$\downarrow$}
& \shortstack{NC$\downarrow$}
& \shortstack{NS$\downarrow$}
& \shortstack{PRWL$\downarrow$}
& \shortstack{RR$\uparrow$}
& \shortstack{\#Vias}
& \shortstack{R-Pass(\%)$\uparrow$}
& \shortstack{Sch.}
& \shortstack{Ref.}
& \shortstack{E2E Pass$\uparrow$} \\
\noalign{\vspace{-2pt}}
\midrule

\rowcolor{gray!20}
\noalign{\vspace{-2.5pt}}
\multicolumn{17}{c}{Commercial Chatbot Systems} \\
\noalign{\vspace{-2.5pt}}
\midrule

\noalign{\vspace{-2.5pt}}
\multirow{5}{*}{GPT-5.5} & \multirow{5}{*}{-}
& Zeroshot
& \best{0.07} & 0.17 & 130.07 & 72.04 & 628.78 & 177.61 & \best{29514.95} & 479.03 & \best{0.90} & 12.47 & 15.52 & {1.189} & {0.521} & 15.23 \\
& & Sch. SW-Net.
& \best{0.08} & \best{0.17} & 150.19 & 64.84 & \second{623.91} & 181.50 & \best{28731.48} & 390.10 & \best{0.92} & 8.98 & \second{52.75} & {0.937} & {0.593} & 34.27 \\
& & Sch. Sem.
& \best{0.08} & \best{0.17}  & 150.56 & 64.07 & \best{619.65} & 178.54 & \best{28141.41} & 432.88 & \second{0.83} & 7.05 & \best{79.56} & 63.71 & {0.578} & {49.08} \\
& & Sch. Img
& \best{0.07} & \best{0.17}  & 144.93 & 72.04 & {628.78} & 177.61 & \best{29514.95} & 435.99 & \second{0.91} & 10.91 & \best{59.55} & 1.120 & {0.559} & 42.83 \\
& & Fewshot
& \best{0.07} &\best{0.19} & 122.46 & \second{99.01} &  \best{745.75} & \best{298.28} & \best{39938.09} & 282.75 & \second{0.95} & {1.00} & \best{63.24} & {1.063} & {0.535} &  \best{62.61}\\

\noalign{\vspace{-2pt}}
\midrule

\noalign{\vspace{-2.5pt}}
\multirow{5}{*}{GPT-5-mini} & \multirow{5}{*}{-}
& Zeroshot
& 0.48 & 0.24 & 87.77 & \second{98.99} & 912.87 & 307.64 & 127023.27 & 271.62 & 0.84 & 6.40 & 41.60 & 1.288 & 0.486 & \second{41.11} \\
& & Sch. SW-Net.
& 0.54 & 0.24 & 70.53 & \second{98.93} & 870.62 & 285.57 & 105211.90 & 195.09 & 0.85 & 4.40 & 44.58 & 0.924 & 0.531 & 44.08 \\
& & Sch. Sem.
& 0.54 & 0.24 & 71.85 & \second{99.29} & 871.02 & 289.87 & 106020.51 & 253.41 & 0.81 & 3.92 & 60.00 & 97.03 & 0.514 & \second{56.22} \\
& & Sch. Img
& 0.48 & 0.24 & 72.48 & \second{98.99} & 912.87 & 307.64 & 127023.27 & 213.99 & 0.79 & 4.87 & 39.56 & {1.148} & 0.494 & 38.91 \\
& & Fewshot
& 0.38 & 0.26 & 66.33 & \best{100.00} & 915.73 & \second{303.91} & 79087.22 & 115.58 & 0.77 & 0.20 & 51.53 & 1.530 & 0.437 & 51.53 \\

\noalign{\vspace{-2pt}}
\midrule

\noalign{\vspace{-2.5pt}}
\multirow{5}{*}{\shortstack{Claude\\Sonnet 4.6}} & \multirow{5}{*}{-}
& Zeroshot
& 0.49 & 0.23 & \second{10.67} & 90.18 & 700.57 & 177.63 & 42531.39 & 245.46 & 0.76 & 5.65 & \second{43.21} & {1.161} & 0.486 & 27.84 \\
& & Sch. SW-Net.
& 0.49 & 0.23 & \second{10.35} & 90.30 & 696.30 & \second{177.49} & \second{44589.04} & 209.44 & 0.86 & 4.32 & 50.30 & 0.923 & {0.578} & \second{45.33} \\
& & Sch. Sem.
& 0.49 & 0.23 & \second{10.38} & 90.30 & 702.26 & \second{177.88} & \second{45734.27} & 284.77 & \second{0.83} & 4.11 & \second{63.19} & 88.34 & 0.548 & 55.56 \\
& & Sch. Img
& 0.49 & 0.23 & \second{10.08} & 90.18 & 700.57 & 177.63 & 42531.39 & 253.40 & 0.85 & 5.04 & 49.90 & 1.041 & 0.539 & \second{44.97} \\
& & Fewshot
& 0.42 & 0.25 & \second{11.47} & \best{100.00} & 939.47 & 313.46 & 69042.98 & 190.43 & 0.54 & 2.43 & 53.10 & 1.315 & 0.487 & 53.10 \\

\noalign{\vspace{-2pt}}
\midrule

\noalign{\vspace{-2.5pt}}
\multirow{5}{*}{\shortstack{Gemini 3.1\\Pro-Preview}} & \multirow{5}{*}{-}
& Zeroshot
& \second{0.26} & \best{0.10} & 116.10 & 42.89 & 672.20 & 199.23 & \second{32692.88} & 419.16 & 0.34 & 1.31 & 38.59 & 1.260 & {0.502} & 16.55 \\
& & Sch. SW-Net.
& \second{0.13} & \second{0.18} & 127.28 & \best{99.64} & 772.24 & 281.62 & 61503.09 & 387.57 &  \second{0.91} & 9.02 & \best{58.32} & 1.275 & 0.552 & \best{58.18} \\
& & Sch. Sem.
& \second{0.16} & \second{0.19} & 129.34 & 99.58 & 747.65 & 268.01 & 52865.44 & 353.77 & 0.63 & 4.75 & 62.64 &{1.102} & 0.552 & \best{62.38} \\
& & Sch. Img
& \second{0.24} & \second{0.20} & 118.15 & \best{99.52} & 658.02 & 207.08 & \second{32931.45} & 316.27 & \best{0.92} & 6.68 & \second{54.94} & 1.031 & {0.577} & \best{54.61} \\
& & Fewshot
& \second{0.16} & 0.22 & 127.48 & \best{100.00} & 836.68 & 332.20 & 56154.95 & \second{97.65} & 0.73 & 0.80 & \second{56.25} & 1.199 & 0.550 & \second{56.25} \\

\noalign{\vspace{-2pt}}
\midrule

\noalign{\vspace{-2.5pt}}
\multirow{5}{*}{\shortstack{Gemini 2.5\\Flash-lite}} & \multirow{5}{*}{-}
& Zeroshot
& 0.61 & 0.32 & \best{6.50} & 89.53 & 1080.87 & 285.58 & 523166.68 & \second{192.76} & 0.20 & 5.12 & 42.89 & 1.165 & 0.486 & 38.40 \\
& & Sch. SW-Net.
& 0.57 & 0.33 & \best{6.57} & 86.67 & 1085.53 & 326.74 & 313629.50 & \best{138.05} & 0.79 & 3.05 & 41.37 & 1.196 & 0.482 & 35.51 \\
& & Sch. Sem.
& 0.58 & 0.33 & \best{6.21} & 86.44 & 1118.50 & 334.70 & 361803.42 & 285.80 & 0.21 & 4.79 & 48.16 & 1.369 & 0.456 & 41.63 \\
& & Sch. Img
& 0.61 & 0.32 & \best{8.43} & 89.53 & 1080.87 & 285.58 & 523166.68 & \second{164.08} & 0.83 & 3.56 & 40.20 & 1.026 & 0.503 & 35.51 \\
& & Fewshot
& 0.51 & 0.38 & \best{4.81} & \best{98.12} & 1354.06 & 332.89 & 180152.54 & 131.82 & 0.62 & 0.50 & 37.83 & 1.530 & 0.452 & 37.12 \\

\noalign{\vspace{-2pt}}
\midrule

\rowcolor{gray!20}
\noalign{\vspace{-2.5pt}}
\multicolumn{17}{c}{Open-Source MLLMs} \\
\noalign{\vspace{-2.5pt}}
\midrule

\noalign{\vspace{-2.5pt}}
\multirow{5}{*}{\shortstack{LLaMA 4\\Maverick}} & \multirow{5}{*}{400B}
& Zeroshot
& 0.54 & 0.28 & 40.67 & \best{99.46} & 1037.73 & 261.07 & 368557.83 & 225.27 & 0.82 & 5.15 & \best{43.60} & 1.494 & 0.478 & \best{43.37} \\
& & Sch. SW-Net.
& 0.53 & 0.27 & 32.32 & \best{99.64} & 1054.92 & 274.45 & 226205.34 & 195.67 & {0.83} & 4.27 & 38.89 & 1.296 & 0.490 & 38.73 \\
& & Sch. Sem.
& 0.54 & 0.27 & 32.91 & \best{99.64} & 990.15 & 269.40 & 384300.14 & \second{251.62} & \second{0.83} & 4.06 & 58.54 & 97.38 & 0.508 & 55.44 \\
& & Sch. Img
& 0.54 & 0.28 & 30.77 & 99.46 & 1037.73 & 261.07 & 368557.83 & 226.32 & 0.83 & 4.87 & 40.47 & 1.244 & 0.507 & 40.15 \\
& & Fewshot
& 0.50 & 0.29 & 27.94 & \best{100.00} & 1080.05 & 326.79 & 109768.57 & \best{15.86} & \best{1.00} & 0.50 & 30.51 & 1.547 & 0.449 & 30.51 \\

\noalign{\vspace{-2pt}}
\midrule

\noalign{\vspace{-2.5pt}}
\multirow{5}{*}{\shortstack{Ministral 3\\14B}} & \multirow{5}{*}{14B}
& Zeroshot
& 0.58 & 0.30 & 17.33 & 99.52 & 1018.66 & 281.81 & 269168.06 & \best{182.98} & \second{0.85} & 4.12 & 34.77 & 1.172 & 0.493 & 34.62 \\
& & Sch. SW-Net.
& 0.59 & 0.29 & 23.79 & 99.35 & 960.08 & 279.51 & 207435.41 & {186.27} & \best{0.84} & 4.21 & 35.97 & 1.115 & 0.504 & 35.69 \\
& & Sch. Sem.
& 0.57 & 0.30 & 25.42 & 99.35 & 946.81 & 274.31 & 163720.97 & 236.36 & 0.83 & 3.72 & 49.55 & 97.09 & 0.495 & 45.69 \\
& & Sch. Img
& 0.58 & 0.30 & 19.23 & 99.52 & 1018.66 & 281.81 & 269168.06 & \best{177.81} & 0.85 & 4.08 & 41.70 & 1.234 & 0.486 & 41.23 \\
& & Fewshot
& 0.52 & 0.36 & 17.24 & \best{100.00} & 1047.79 & 319.97 & 117151.31 & 548.91 & \best{1.00} & 2.00 & 52.48 & 1.350 & 0.474 & 52.48 \\

\noalign{\vspace{-2pt}}
\midrule

\noalign{\vspace{-2.5pt}}
\multirow{5}{*}{Qwen3.5-9B} & \multirow{5}{*}{9B}
& Zeroshot
& 0.59 & 0.26 & 460.47 & 1.61 & \best{460.30} &  \best{90.59} & 28168.13 & 279.67 & 0.31 & 6.00 & 29.17 & 1.520 & 0.402 & 0.47 \\
& & Sch. SW-Net.
& 0.64 & 0.24 & 598.24 & 5.77 & 891.03 & 247.08 & 70491.50 & 208.04 & 0.77 & 2.92 & 36.36 & 1.318 & {0.576} & 2.14 \\
& & Sch. Sem.
& 0.64 & 0.22 & 380.75 & 6.96 & 912.23 & 269.19 & 92222.56 & 255.94 & 0.84 & 2.92 & 35.45 & 6.96 & 0.521 & 2.32 \\
& & Sch. Img
& 0.60 & 0.27 & 574.55 & 5.18 & \best{591.87} & \best{141.90} & 51896.65 & 115.39 & 0.84 & 3.22 & 37.93 & 1.134 & 0.492 & 1.96 \\
& & Fewshot
& 0.43 & 0.48 & 389.54 & 14.52 & 1646.13 & 934.79 & 186513.73 & 174.26 & 0.81 & 0.33 & 10.46 & 1.245 & 0.449 & 1.52 \\

\noalign{\vspace{-2pt}}
\midrule

\noalign{\vspace{-2.5pt}}
\multirow{5}{*}{\shortstack{Qwen3.6-35B-\\A3B}} & \multirow{5}{*}{35B}
& Zeroshot
& 0.64 & \second{0.13} & 456.81 & 13.21 & \second{603.41} & \second{159.19} & 42572.41 & 250.81 & 0.31 & 1.50 & 40.22 & {1.164} & 0.516 & 5.31 \\
& & Sch. SW-Net.
& 0.62 & 0.21 & 343.63 & 51.22 & \best{594.01} & \best{169.56} & 46996.45 & \second{147.36} & 0.81 & 3.83 & 45.07 & {0.642} & 0.571 & 22.84 \\
& & Sch. Sem.
& 0.62 & 0.20 & 286.55 & 48.07 & \second{634.59} & \best{160.86} & 154873.07 & \best{186.55} & 0.80 & 3.00 & 59.33 & 46.76 & 0.522 & 27.42 \\
& & Sch. Img
& 0.61 & 0.22 & 255.28 & 51.04 & \second{620.16} & \second{174.54} & 70062.54 & 202.15 & 0.83 & 3.79 & 48.24 & {0.981} & 0.477 & 24.45 \\
& & Fewshot
& 0.35 & \second{0.21} & 324.74 & 92.33 & \second{767.92} & 310.12 & \second{56298.48} & 462.58 & 0.76 & 1.50 & 44.75 & {1.108} & {0.537} & 41.32 \\

\noalign{\vspace{-2pt}}
\midrule

\rowcolor{gray!20}
\noalign{\vspace{-2.5pt}}
\multicolumn{17}{c}{Commercial PnR Tools} \\
\noalign{\vspace{-2pt}}
\midrule

\noalign{\vspace{-1.5pt}}
Quilter~\cite{quilter2026} & - & - & 0.023 & 0.019 & 5841.00 & -- & 553.48 & 46.37 & 1.65 & 194.28 & 0.84 & 163.27 & -- & 1.24 & 0.57 & -- \\

\noalign{\vspace{-1.5pt}}
\midrule

\rowcolor{gray!20}
\noalign{\vspace{-2pt}}
\multicolumn{17}{c}{Open-Source PnR Tools} \\
\noalign{\vspace{-2pt}}
\midrule

\noalign{\vspace{-1.5pt}}
PCBAgent~\cite{chen2025pcbagent} & - & - & 0.41 & 0.24 & 316.57 & -- & 1077.03 & 372.44 & 109405.37 & 244.79 & 0.42 & 5.51 & -- & 0.30 & 0.41 & -- \\

\noalign{\vspace{-1.5pt}}
\midrule

\noalign{\vspace{-1.5pt}}
SA-PCB~\cite{sapcb2024} & - & - & 0.22 & 0.28 & 40.94 & -- & 1098.81 & 334.60 & 129177.56 & 165.09 & 0.25 & 3.44 & -- & 4.22 & 0.39 & -- \\

\noalign{\vspace{-1.5pt}}
\midrule

\noalign{\vspace{-1.5pt}}
NS-Place~\cite{cheng2022net} & - & - & 0.73 & 0.23 & 10.87 & -- & 1010.54 & 295.80 & 89299.07 & 168.45 & 0.23 & 3.85 & -- & 1.84 & 0.38 & -- \\

\noalign{\vspace{-1.5pt}}
\midrule

\noalign{\vspace{-1.5pt}}
Cypress~\cite{zhang2025cypress} & - & - & 0.24 & 0.18 & 26.83 & -- & 994.54 & 332.59 & 128294.67 & 300.61 & 0.33 & 4.90 & -- & 4.68 & 0.49 & -- \\

\noalign{\vspace{-1.5pt}}
\midrule

\noalign{\vspace{-1.5pt}}
Chen et al.~\cite{chen2025clearance} & - & - & 0.19 & 0.20 & 131.37 & -- & 602.21 & 130.03 & 22950.45 & 369.99 & 0.62 & 6.11 & -- & 3.19 & 0.86 & -- \\

\noalign{\vspace{-1.5pt}}
\midrule

\noalign{\vspace{-1.5pt}}
Human Engineer (Ref. Layout) & - & - & 0.01 & 0.14 & -- & 100.00 & 579.91 & 114.98 & 19664.86 & 140.89 & 1.00 & 15.4 & -- & 1.01 & 1.00 & -- \\

\noalign{\vspace{-1.5pt}}
\bottomrule

\end{tabular}%
}

\endgroup
\vspace{-15pt}
\end{table*}

\textbf{Implementation of Commercial and Open-Source PCB 
Layout Baselines.}
To comprehensively evaluate the difficulty and quality of 
our benchmark, we assess placement performance across both 
commercial and academic baselines. For commercial tool, we 
evaluate Quilter~\cite{quilter2026}, a
reinforcement learning-based PCB PnR solution. For academic 
baselines, we implement SA-PCB~\cite{sapcb2024}, an 
OpenROAD-based simulated annealing placer; 
NS-Place~\cite{cheng2022net}, a net separation-oriented 
analytical placement method; Cypress~\cite{zhang2025cypress}, 
a VLSI-inspired GPU-accelerated placement framework; 
Chen et al.~\cite{chen2025clearance}, a clearance-constrained 
global placement method supporting heterogeneous components;
and PCBAgent~\cite{chen2025pcbagent}, an LLM and 
reinforcement learning-based placement framework.

\textbf{Implementation of LMMs Agentic Framework.} 
As shown in Figure \ref{fig:react_framework}, we design a multi-modal, multi-agent, and multi-round agentic evaluation framework where 
LMMs are provided with the board boundary, netlist, component 
dimensions, and schematic diagram, and equipped with three 
tools: layout visualization for rendering the current 
placement state, overlap checking for reporting constraint 
violations, and routability verification for estimating 
routing conflicts. At each step, the LMM observes the 
rendered layout and iteratively refines component positions, 
orientations, and layer assignments until all constraints 
are satisfied or the maximum number of steps is reached.
Agentic metrics are computed only over successfully parsed placements.
API failures and empty or invalid structured responses are reported as
unavailable rather than assigned zero-valued metrics.

\begin{table*}[!ht]
\centering
\caption{\textbf{Agentic PCB component-placement evaluation.}
Lower is better except for RR and Ref. Sim.; NS is reported as a raw value,
and Steps is the mean number of refinement steps.}
\vspace{-0.05in}
\label{tab:agentic}

\begingroup
\tiny
\setlength{\tabcolsep}{4pt}
\renewcommand{\arraystretch}{0.4}

\resizebox{\textwidth}{!}{%
\begin{tabular}{l c c ccc cccccc cc c}
\toprule

\noalign{\vspace{-2pt}}
\multirow{2}{*}{Model}
& \multirow{2}{*}{Size}
& \multirow{2}{*}{\shortstack{Tool\\Setting}}
& \multicolumn{3}{c}{Geometric Legality}
& \multicolumn{6}{c}{Routability}
& \multicolumn{2}{c}{Relational Placement}
& \multicolumn{1}{c}{Agentic Process} \\

\cmidrule(lr){4-6}
\cmidrule(lr){7-12}
\cmidrule(lr){13-14}
\cmidrule(lr){15-15}

& &
& \makecell[c]{OO$\downarrow$}
& \makecell[c]{OoB$\downarrow$}
& \makecell[c]{RT$\downarrow$}
& \makecell[c]{HPWL$\downarrow$}
& \makecell[c]{NC$\downarrow$}
& \makecell[c]{NS$\downarrow$}
& \makecell[c]{PRWL$\downarrow$}
& \makecell[c]{RR$\uparrow$}
& \makecell[c]{\#Vias$\downarrow$}
& \makecell[c]{Sch. Rel.}
& \makecell[c]{Ref. Sim.}
& \makecell[c]{Steps} \\
\noalign{\vspace{-2pt}}
\midrule

\rowcolor{gray!20}
\noalign{\vspace{-2.5pt}}
\multicolumn{15}{c}{Commercial Chatbot Systems} \\
\noalign{\vspace{-2.5pt}}
\midrule

\noalign{\vspace{-2.5pt}}
\multirow{4}{*}{GPT-5.5} & \multirow{4}{*}{-}
& Visualization
& \best{0.015} & \second{0.010} & 410.276 & 622.557 & 167.600 & 15310.159 & \second{113.94} & \second{0.591} & \best{0.00} & 0.966 & 0.586 & 3.900 \\
& & Geo. Check
& 0.003 & \second{0.007} & 295.334 & 790.055 & 357.200 & 35016.412 & \second{201.28} & \second{0.631} & 8.00 & 0.893 & 0.538 & 2.200 \\
& & Routing Score
& 0.010 & \best{0.000} & 517.918 & \best{281.964} & \best{82.000} & \best{2846.929} & 163.17 & \best{0.735} & \best{2.00} & 1.000 & 0.552 & 4.750 \\
& & All Tools
& 0.005 & 0.161 & \second{496.705} & 304.824 & \second{53.375} & 3350.575 & 551.83 & \second{0.509} & \second{2.00} & 0.675 & 0.569 & 5.875 \\
\noalign{\vspace{-2pt}}
\midrule

\noalign{\vspace{-2.5pt}}
\multirow{4}{*}{GPT-5-mini} & \multirow{4}{*}{-}
& Visualization
& \second{0.028} & 0.045 & \best{120.336} & 664.691 & 179.400 & 17363.408 & 387.20 & 0.370 & 2.00 & 1.336 & 0.499 & 2.600 \\
& & Geo. Check
& 0.049 & 0.075 & 813.394 & 1039.503 & 522.200 & 67860.676 & 222.05 & 0.347 & 7.00 & 1.569 & 0.465 & 16.800 \\
& & Routing Score
& 0.095 & 0.046 & 522.033 & 824.905 & 392.800 & 49582.308 & 530.78 & 0.200 & 8.00 & 1.633 & 0.429 & 12.400 \\
& & All Tools
& 0.024 & 0.293 & 1362.921 & \second{241.864} & 59.500 & \second{2722.143} & 204.42 & 0.300 & 5.00 & 1.327 & 0.400 & 28.750 \\
\noalign{\vspace{-2pt}}
\midrule

\noalign{\vspace{-2.5pt}}
\multirow{4}{*}{\shortstack{Claude\\Sonnet 4.6}} & \multirow{4}{*}{-}
& Visualization
& 0.029 & \best{0.000} & 120.738 & \second{288.944} & 69.714 & \second{3775.568} & 132.57 & 0.540 & 3.67 & 1.344 & 0.553 & 4.571 \\
& & Geo. Check
& 0.011 & 0.028 & 546.897 & 790.849 & 330.400 & 30869.894 & 380.68 & 0.506 & 7.00 & 1.029 & 0.515 & 11.000 \\
& & Routing Score
& 0.055 & 0.014 & 509.824 & 835.031 & 421.000 & 60252.539 & 270.42 & \second{0.614} & \best{2.00} & 1.093 & 0.519 & 14.000 \\
& & All Tools
& 0.004 & \second{0.106} & 854.905 & 455.474 & 97.625 & 11560.411 & 343.92 & \best{0.638} & 5.50 & 1.829 & 0.411 & 22.625 \\
\noalign{\vspace{-2pt}}
\midrule

\noalign{\vspace{-2.5pt}}
\multirow{4}{*}{\shortstack{Gemini 3.1\\Pro-Preview}} & \multirow{4}{*}{-}
& Visualization
& 0.071 & 0.023 & 125.666 & 306.256 & \second{69.000} & 4350.194 & \best{84.35} & 0.586 & \best{0.00} & 0.920 & 0.713 & 1.750 \\
& & Geo. Check
& 0.007 & 0.030 & 417.642 & 1068.203 & 506.400 & 77213.345 & 347.12 & 0.484 & \second{5.00} & 0.852 & 0.553 & 3.400 \\
& & Routing Score
& 0.015 & 0.033 & \second{441.500} & 1145.583 & 616.667 & 65857.878 & \best{112.59} & 0.480 & 4.3 & 0.942 & 0.525 & 3.000 \\
& & All Tools
& 0.009 & 0.349 & 536.814 & 577.776 & 117.143 & 13879.121 & 50.45 & 0.455 & \best{1.00} & 1.070 & 0.520 & 5.000 \\
\noalign{\vspace{-2pt}}
\midrule

\noalign{\vspace{-2.5pt}}
\multirow{4}{*}{\shortstack{Gemini 2.5\\Flash}} & \multirow{4}{*}{-}
& Visualization
& 0.049 & 0.441 & 2138.082 & \best{147.931} & \best{18.000} & \best{1666.811} & 147.81 & \best{0.667} & 4.00 & 1.372 & 0.404 & 30.000 \\
& & Geo. Check
    & 0.098 & \best{0.000} & \best{29.648} & \best{306.526} & \best{102.500} & \best{3383.558} & 197.88 & \best{0.677} & 7.00 & 0.957 & 0.505 & 1.000 \\
& & Routing Score
& 0.082 & \best{0.000} & 864.272 & 264.859 & 88.500 & \second{3662.414} & \second{130.44} & 0.274 & 5.00 & 2.084 & 0.406 & 20.000 \\
& & All Tools
& 0.000 & \best{0.000} & \best{7.802} & \best{121.400} & \best{28.000} & \best{965.265} & \second{49.83} & 0.172 & 6.87 & 1.137 & 0.760 & 3.000 \\
\noalign{\vspace{-2pt}}
\midrule

\rowcolor{gray!20}
\noalign{\vspace{-2.5pt}}
\multicolumn{15}{c}{Open-Source MLLMs} \\
\noalign{\vspace{-2.5pt}}
\midrule

\noalign{\vspace{-2.5pt}}
\multirow{4}{*}{\shortstack{LLaMA 4\\Maverick}} & \multirow{4}{*}{400B}
& Visualization
& 0.121 & 0.141 & \second{120.453} & 373.654 & 117.400 & 6369.738 & 142.07 & 0.248 & 4.33 & 1.413 & 0.527 & 5.000 \\
& & Geo. Check
& 0.104 & 0.061 & 579.333 & \second{393.620} & \second{108.000} & \second{7751.014} & \best{108.29} & 0.599 & \best{3.00} & 2.308 & 0.362 & 19.667 \\
& & Routing Score
& 0.107 & \second{0.050} & 831.474 & 1000.486 & 558.200 & 80826.012 & 175.97 & 0.562 & 4.50 & 2.282 & 0.371 & 19.800 \\
& & All Tools
& 0.051 & 0.142 & 854.743 & 409.496 & 95.500 & 8014.048 & \best{43.92} & 0.287 & 1.00 & 0.986 & 0.545 & 50.000 \\
\noalign{\vspace{-2pt}}
\midrule

\noalign{\vspace{-2.5pt}}
\multirow{4}{*}{\shortstack{Ministral\\14B}} & \multirow{4}{*}{14B}
& Visualization
& 0.130 & 0.053 & 84.293 & 691.764 & 151.429 & 28092.990 & 126.68 & 0.252 & \second{0.50} & 1.237 & 0.470 & 3.286 \\
& & Geo. Check
& 0.177 & 0.145 & \second{249.803} & 881.112 & 442.750 & 34241.999 & 541.25 & 0.167 & 8.00 & 2.056 & 0.418 & 9.750 \\
& & Routing Score
& 0.124 & 0.101 & \best{169.451} & \second{370.257} & \second{122.750} & 7451.601 & 179.62 & 0.530 & \second{2.50} & 1.576 & 0.421 & 8.250 \\
& & All Tools
& 0.065 & 0.180 & 1515.420 & 865.037 & 214.667 & 34773.643 & 276.47 & 0.449 & 3.00 & 2.012 & 0.377 & 50.000 \\
\noalign{\vspace{-2pt}}
\midrule

\noalign{\vspace{-2.5pt}}
\multirow{4}{*}{Qwen3.5-9B} & \multirow{4}{*}{9B}
& Visualization
& -- & -- & -- & -- & -- & -- & -- & -- & -- & -- & -- & -- \\
& & Geo. Check
& -- & -- & -- & -- & -- & -- & -- & -- & -- & -- & -- & -- \\
& & Routing Score
& -- & -- & -- & -- & -- & -- & -- & -- & -- & -- & -- & -- \\
& & All Tools
& -- & -- & -- & -- & -- & -- & -- & -- & -- & -- & -- & -- \\
\noalign{\vspace{-2pt}}
\midrule

\noalign{\vspace{-2.5pt}}
\multirow{4}{*}{\shortstack{Qwen3.6-35B-\\A3B}} & \multirow{4}{*}{35B}
& Visualization
& -- & -- & -- & -- & -- & -- & -- & -- & -- & -- & -- & -- \\
& & Geo. Check
& -- & -- & -- & -- & -- & -- & -- & -- & -- & -- & -- & -- \\
& & Routing Score
& -- & -- & -- & -- & -- & -- & -- & -- & -- & -- & -- & -- \\
& & All Tools
& -- & -- & -- & -- & -- & -- & -- & -- & -- & -- & -- & -- \\
\noalign{\vspace{-2pt}}
\midrule

\noalign{\vspace{-2pt}}
\bottomrule
\end{tabular}%
}

\vspace{2pt}
\parbox{\textwidth} {\tiny
    Qwen3.5-9B and Qwen3.6-35B-A3B tool-setting agentic results are unavailable.
The supplied placement archive contains unlabelled one-pass Qwen outputs, but
does not identify them as Visualization, Geo. Check, Routing Score, or All
Tools runs. We therefore report these entries as ``--'' and exclude them from
aggregate calculations; ``--'' does not denote zero performance. Layout
metrics, RT, and Steps are averaged over finalized graph JSON outputs. PRWL and
\#Vias are reported over fully routed instances, while RR assigns zero to
failed routing attempts. The Gemini 2.5 Visualization row uses Gemini 2.5 Flash, while its All Tools row uses Gemini 2.5 Flash-lite and has one finalized placement. Its packaged routing summary does not provide replacement PRWL or \#Vias values, so the original table values are retained for those two cells.
}

\endgroup
\vspace{-15pt}
\end{table*}

\vspace{-0.1in}
\subsection{Main Results and Findings}
\vspace{-0.1in}
\textbf{Evaluation results of zero-shot study and agentic study} are shown in Table~\ref{tab:zeroshot} and Table \ref{tab:agentic}. Implementation details are provided in \tai{Appendix \ref{sec:appendix:exp}} due to space constraints. Overall, the dense component placement performance of LMMs remains significantly below that of human engineers, highlighting their limited capability in comprehensive layout design. Evaluation results reveal consistent limitations across multiple layout design capabilities.

\textbf{$\bullet$ Geometric Legality.} 
While LMMs demonstrate promising performance on simple PCB layouts with fewer than 40 components per design, their performance degrades significantly as complexity and component density increase. 
Furthermore, LMMs exhibit poor spatial utilization, placing \tai{80\%} of components on the top layer rather than distributing them across available routing layers, suggesting a limited understanding of three-dimensional spatial organization in PCB design.

\textbf{$\bullet$ Routability.}
LMMs exhibit substantial weakness in routability-aware placement, treating component arrangement as a purely spatial packing problem while neglecting circuit connectivity. 
Although zeroshot LMMs achieve promising score in overlap of less than \tai{10\%} and out of boundary of less than \tai{10\%}, their placements consistently result in poor routing outcomes reflected in higher PRWL and lower RR, suggesting a fundamental gap in understanding the relationship between component placement and downstream routing feasibility.

\textbf{$\bullet$ Relational Preservation.} 
LMMs achieve a similarity score below 60\% when comparing predicted placements against reference layout designs, indicating that component positions deviate substantially from both the schematic diagram's topology and the expert-designed reference in terms of relative spatial distances. This quantitative gap reflects a deficiency: LMMs lack a structured understanding of circuit hierarchy and functional grouping, and consequently fail to preserve the spatial relationships of subcircuits during placement.


\textbf{$\bullet$ Agentic Tool Use.}
Agentic placement is practically important because it allows models
to inspect and iteratively repair intermediate layouts. Tool
effectiveness is nevertheless model- and metric-dependent. Among the
evaluated feedback settings, the All Tools condition yields the lowest
overlap error for Claude-Sonnet-4.6, Ministral-14B, and GPT-5-mini
(OO $=0.004$, $0.065$, and $0.024$, respectively). However, All Tools
does not consistently achieve the best performance on other metrics,
such as out-of-boundary violations, routability, and relational
agreement. These results indicate that combining multiple feedback
channels can improve overlap correction, while effective agentic
placement still requires model- and objective-aware tool selection.
Additional analysis is provided in
Appendix~\ref{sec:agentic_analysis}.

\textbf{$\bullet$ Agentic Spatial \& Routing Reasoning.} 
By incorporating layout visualization, legality checking, and routability feedback into a multi-agent framework, agentic interaction improves placement performance by reducing component overlap and out-of-boundary violations, while layout visualization feedback contributes to improved routing performance compared to direct generation baselines.


\begin{figure*}[t]
\centering
\includegraphics[
    width=\textwidth,
    keepaspectratio
]{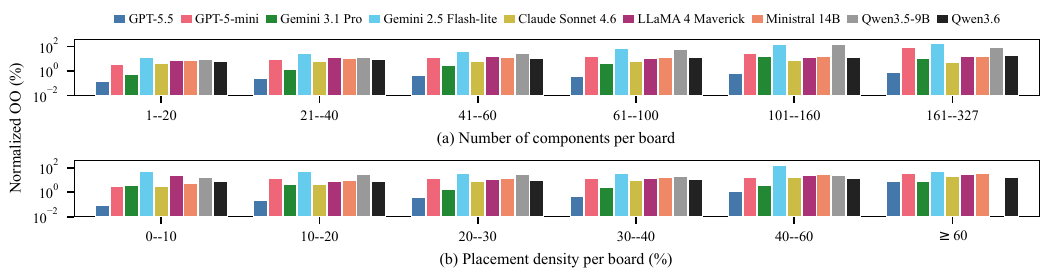}
\caption{
LMM placement performance under increasing PCB complexity:
(a) component count per board and (b) placement density.
Lower overlap is better.
}
\label{fig:diagnostic_quartiles}
\vspace{-0.2in}
\end{figure*}


\subsection{Diagnostic Analyses}
\label{sec:diagnostic_analysis}

We further conduct additional diagnostic analyses
covering constraint-violation frequency, performance across board
complexity and density, the contributions of different input
representations and feedback channels, and recurring PCB design
principles missed by current models.

\textbf{$\bullet$ Most frequently violated constraints.}
Across the successfully parsed LMM outputs included in this diagnostic,
{79.82\%} contain at least one inter-component pad or SMD
overlap, making overlap the most frequent evaluated geometric
violation. Boundary-containment violations are similarly widespread:
{72.89\%} of placements contain at least one component block,
pad, SMD, or hole extending outside the valid board boundary. These
results show that the aggregate OO and OoB scores reflect recurring
constraint failures rather than a small number of extreme cases.

\textbf{$\bullet$ Impact of board complexity and density.}
As shown in Figure~\ref{fig:diagnostic_quartiles}, we analyze placement performance across boards with component counts from 1 to 327 and placement densities from $0.49\%$ to $90.48\%$. Complexity is measured by component count, while placement density is the total pad, SMD, and hole area divided by valid board area. As either factor increases, normalized overlap area (OO; lower is better) also increases, with median relative increases of $504\%$ for component count and $337.8\%$ for placement density respectively. Overall, higher board complexity leads to more component overlap and worse placement performance.

\textbf{$\bullet$ Contributions of one-pass input representations.}
The effectiveness of additional schematic information depends on both
the model and the evaluation dimension. For functional grouping, all
15 comparisons across five models and three schematic representations
improve over their matched no-schematic conditions. The mean paired AUC
gains are +0.079 for the schematic spatially weighted netlist,
+0.089 for the schematic image, and +0.091 for
semantic component attributes. The best schematic-aware
functional-grouping AUC is 0.738 for Gemini 3.1 Pro, 0.727 for
Claude Sonnet 4.6, 0.695 for LLaMA 4 Maverick, 0.669 for
Ministral 14B, and 0.657 for GPT-5-mini, compared with
0.590--0.646 without schematic context. In contrast, few-shot
prompting changes mean AUC by $-0.006$ and improves only 42.8\% of
paired-board comparisons. Thus, no input representation is uniformly
best across models and metrics.

\textbf{$\bullet$ Contributions of agentic feedback channels.}
Feedback effectiveness is similarly model- and objective-dependent.
As shown in Table~\ref{tab:agentic}, Claude Sonnet 4.6 and
Ministral 14B achieve their lowest overlap under Geo.\ Check, where the
model receives explicit identities and locations of conflicting
components. This structured textual feedback removes the need to infer
overlap locations from the rendered image and supports localized
corrections. In contrast, GPT-5-mini achieves its lowest overlap under
Routing Score, suggesting that scalar HPWL feedback can also guide
useful geometric revisions for some models. Combining all feedback
channels is therefore not uniformly optimal.

\textbf{$\bullet$ Systematically missed PCB design principles.}
The diagnostic analyses reveal four recurring weaknesses:
\textbf{(i)} maintaining pad-level non-overlap,
\textbf{(ii)} keeping all component geometry within the board
boundary,
\textbf{(iii)} preserving proximity among electrically connected or
schematically co-grouped components, and
\textbf{(iv)} adapting placement strategies to crowded or
component-heavy boards. Frontier LMMs occasionally capture these
principles, but they do not apply them reliably across boards, input
representations, or feedback conditions.

\section{Conclusion}
We introduce \textit{\name}, a benchmark for evaluating LMMs on PCB layout geometric reasoning, consisting of \tai{1,681} real-world PCB layouts coupled with fine-grained annotations and structured evaluation protocols. We evaluate state-of-the-art LMMs under both end-to-end and tool-augmented settings, alongside open-source and commercial PCB placement tools as baselines. 
Our analysis reveals persistent limitations in geometric legality,
routability optimization, agentic refinement, and preservation of
schematic-derived relationships.

\newpage
\section*{Limitations}\label{limitation}
The agentic evaluation does not contain more than 50 turn iterations, primarily due to the substantial computational cost and resource requirements of iterative agentic inference, and thus may miss evaluating placement refinement scenarios that require longer iterative convergence.

We evaluate each of the three tools (layout visualization, overlap checking, and routability verification) in isolation as well as in full combination to assess their individual and collective impact on placement quality. 
The benchmark does not cover all possible tool combination settings for agentic evaluation; for example, partial subsets such as pairing layout visualization with overlap checking are not explored in this work. 
It is an extensive research direction and beyond the scope of this work. 
However, as the benchmark is open-source, we plan to extend it to include broader agentic evaluation settings in the future.

This work focuses on spatial reasoning for PCB component placement, and thus does not study the routing performance of LMMs directly. Instead, we employ FreeRouting as a fixed backbone to evaluate the quality of placements produced by LMMs, isolating placement as the primary variable of interest. Routing optimization remains an important and complementary research direction that we leave for future work.

\section*{Ethics Statement}\label{ethics}

\textbf{Intended usage.} We aim for the benchmark to serve as a comprehensive evaluation platform for assessing both frontier commercial and open-source LMMs and state-of-the-art commercial and academic PnR tools on real-world PCB layout design at scale. The code and dataset are publicly released under the CC-BY-4.0 license, consistent with the licensing terms of all constituent datasets incorporated in the benchmark.

\noindent\textbf{Accessibility and Potential Misuse.} The main goal of our work is to encourage the community to use the benchmark to advance PCB layout design research and to highlight the current limitations of existing LMMs and PnR tools on real-world placement challenges. Our evaluations reveal that existing LMMs achieve notably poor performance on PCB layout design tasks, underscoring a significant gap between current model capabilities and the demands of real-world PCB design. However, as the dataset is primarily sourced from DIY and open-source designs, unknown biases inherited from these sources may influence evaluation outcomes and limit the benchmark's coverage of more advanced PCB designs. We tried to mitigate this by incorporating diverse PCB designs from various sources, and we encourage the community to use this benchmark as a foundation to drive future research toward developing models better suited for practical PCB layout design.

\noindent\textbf{Ai Assistants.}
We used AI-assisted tools to support proofreading and improve the clarity, grammar, and readability of the manuscript. All technical content, analysis, and final revisions were carefully reviewed and validated by the authors.

\section{Acknowledgments}
This research was supported by the U.S. National Science Foundation (NSF) under CAREER Award No. 2046972. We also gratefully acknowledge support from the Google Gemini Academic Program, which provided access to Gemini models.

\bibliography{anthology}
\newpage

\appendix

\twocolumn[{%
  \begin{center}
    \vspace{-0.1in}
    \includegraphics[width=1.\textwidth]{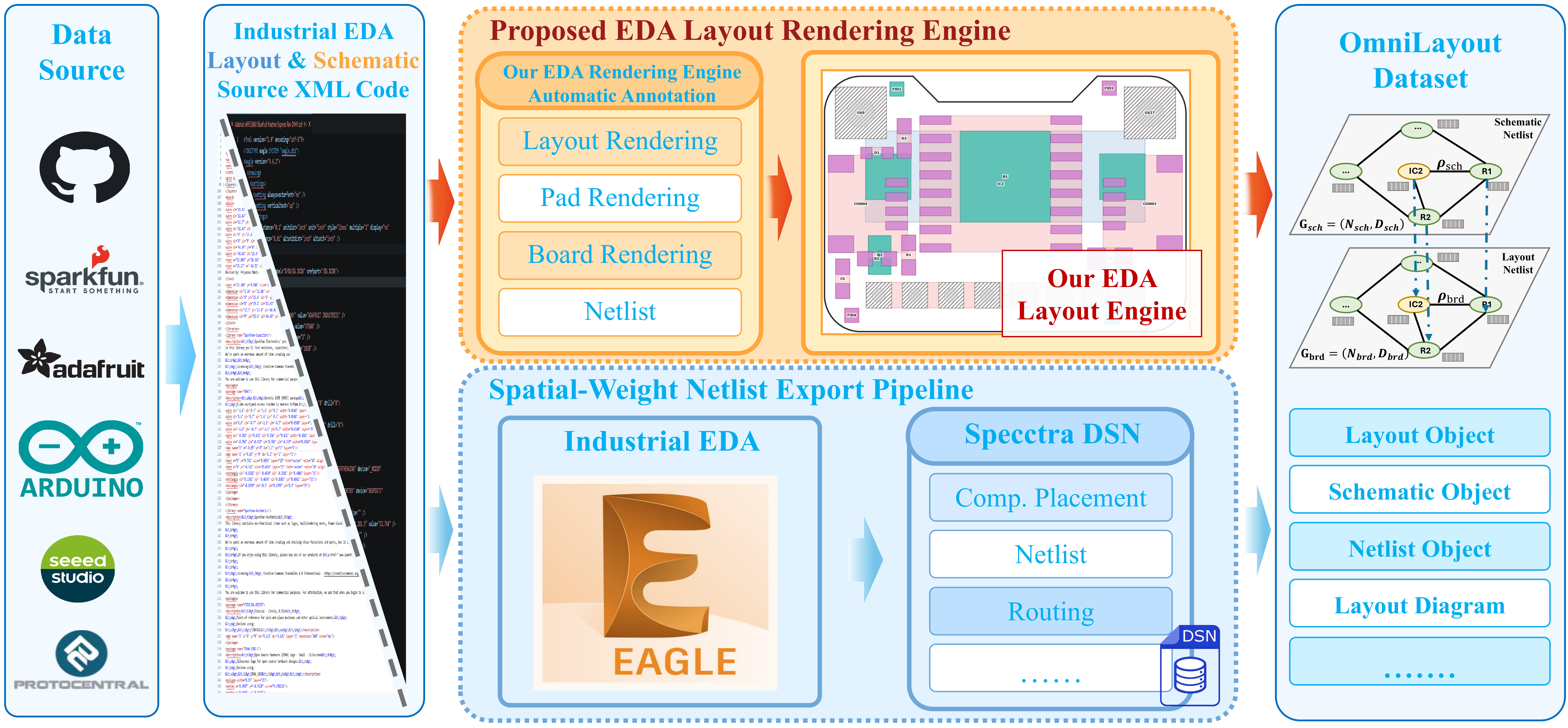}
    \captionof{figure}{\textbf{Automated PCB layout annotation pipeline for {\name}.} We directly export Specctra DSN files from industrial EDA tools and leverage our EDA layout rendering engine to automatically generate PCB layout annotations, including component placements, orientations, board geometry, routing information, and spatial-aware netlists.}
    \label{fig:annotation_pipeline}
  \end{center}
}]

\section{Data Collection}\label{sec:appendix:data}

\begin{figure*}[h!]
  \begin{minipage}[t]{0.44\linewidth}
    \centering
    \includegraphics[width=\linewidth]{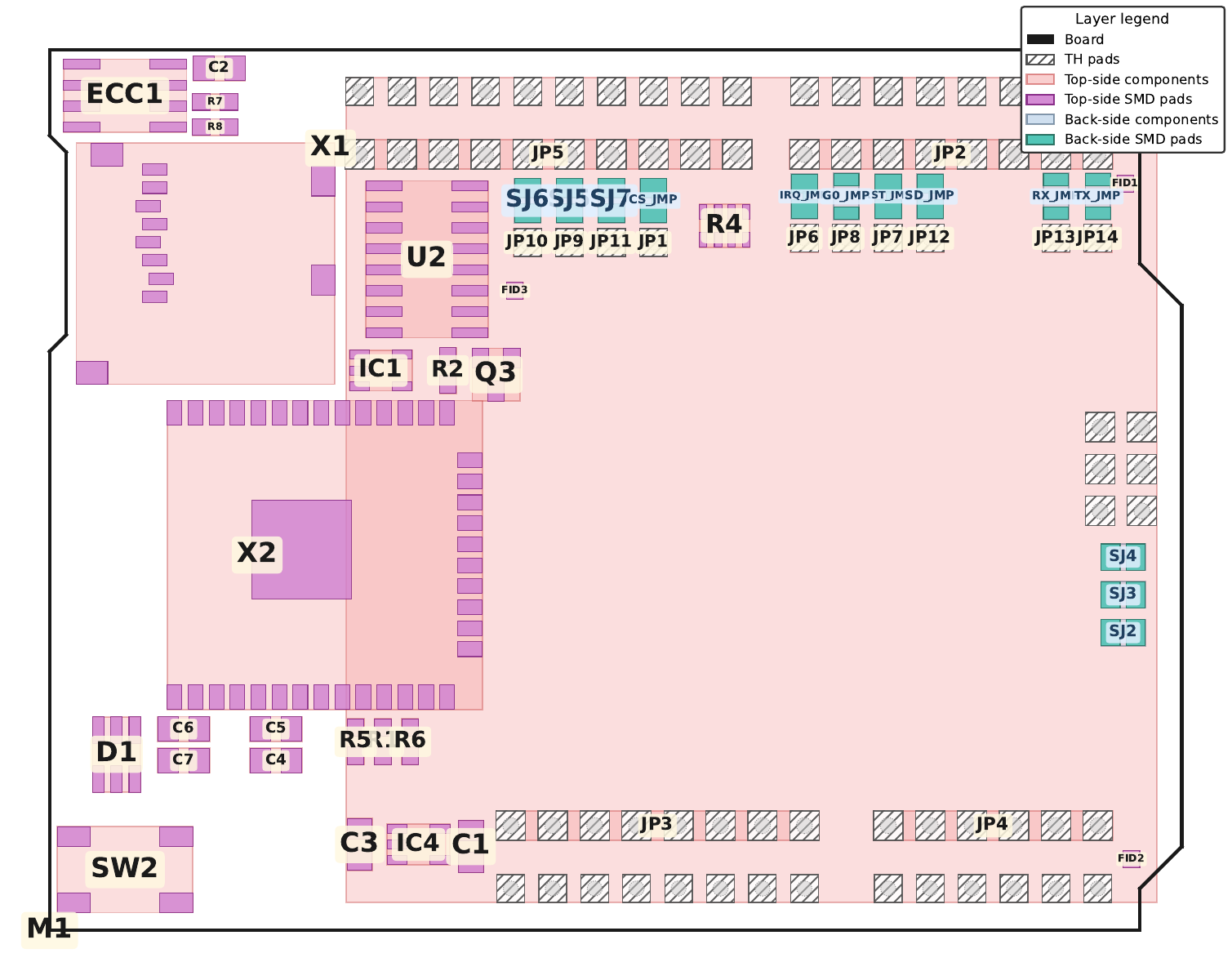}
    \subcaption{Layout Diagram by Custom EDA Layout Engine}
  \end{minipage} \hfill
  \begin{minipage}[t]{0.44\linewidth}
    \centering
    \includegraphics[width=\linewidth]{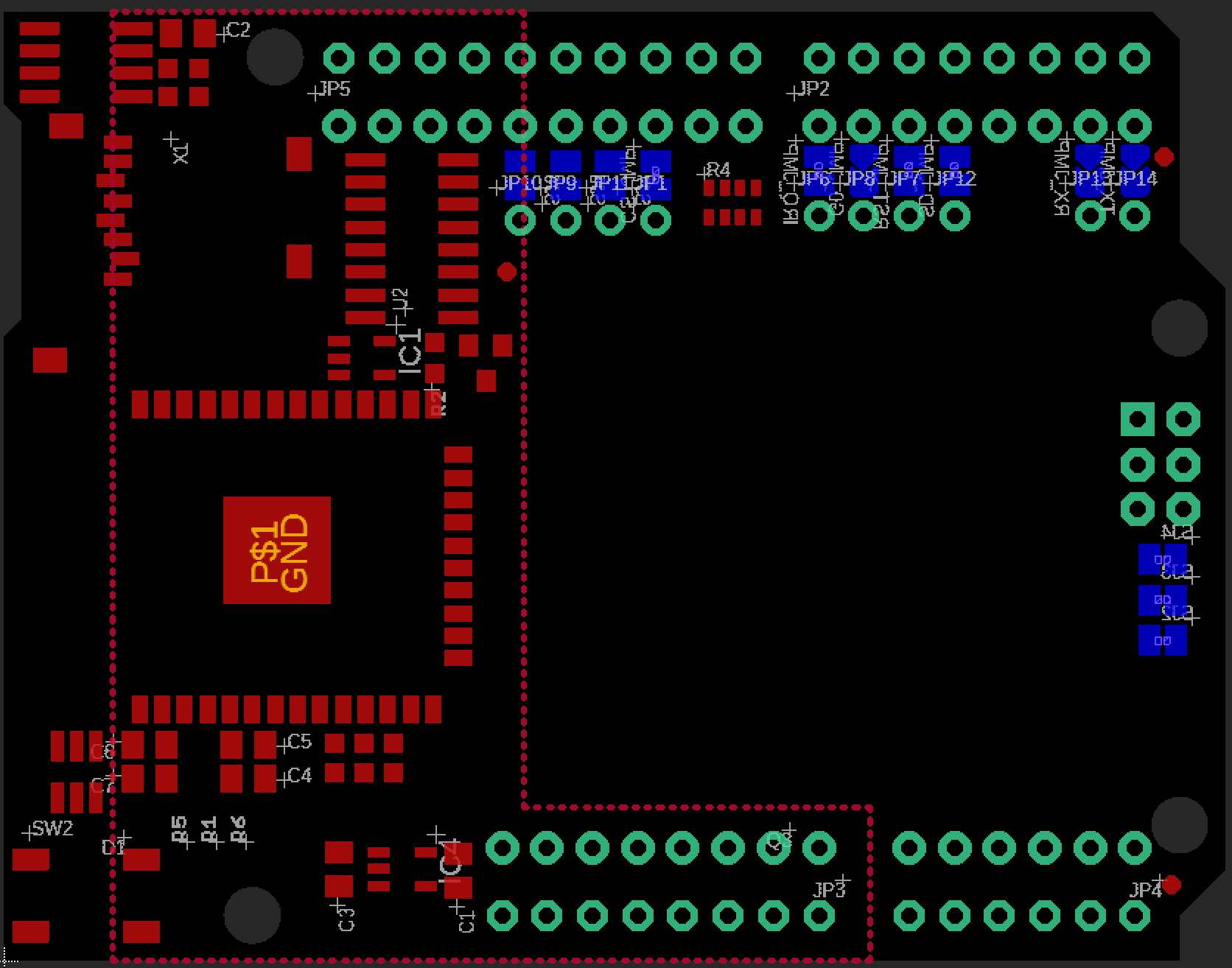}
    \subcaption{Layout Diagram by Industrial EDA Layout Engine}
  \end{minipage} \\[6pt]

  \begin{minipage}[t]{0.44\linewidth}
    \centering
    \includegraphics[width=\linewidth]{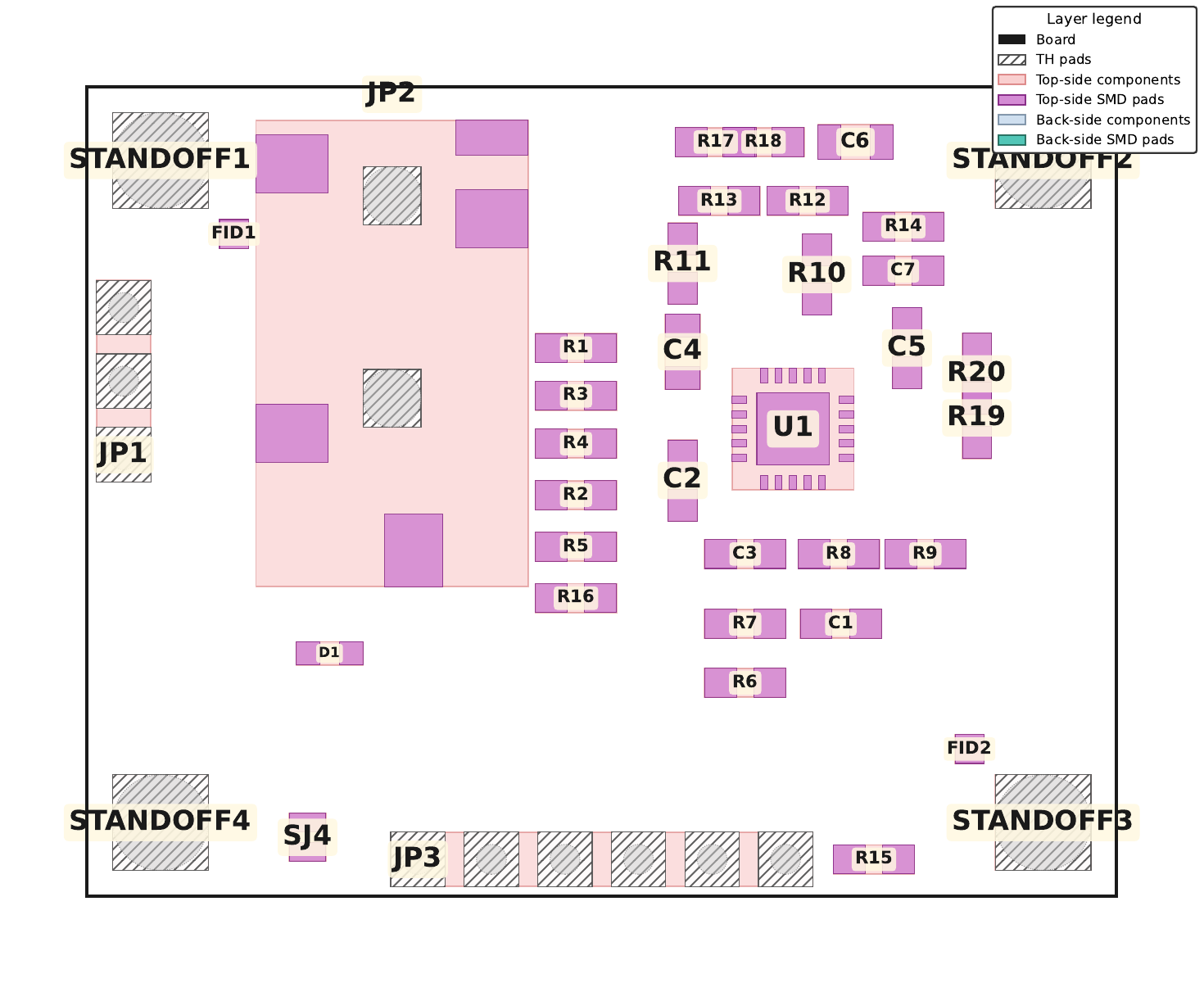}
    \subcaption{Layout Diagram by Custom EDA Layout Engine}
  \end{minipage} \hfill
  \begin{minipage}[t]{0.44\linewidth}
    \centering
    \includegraphics[width=\linewidth]{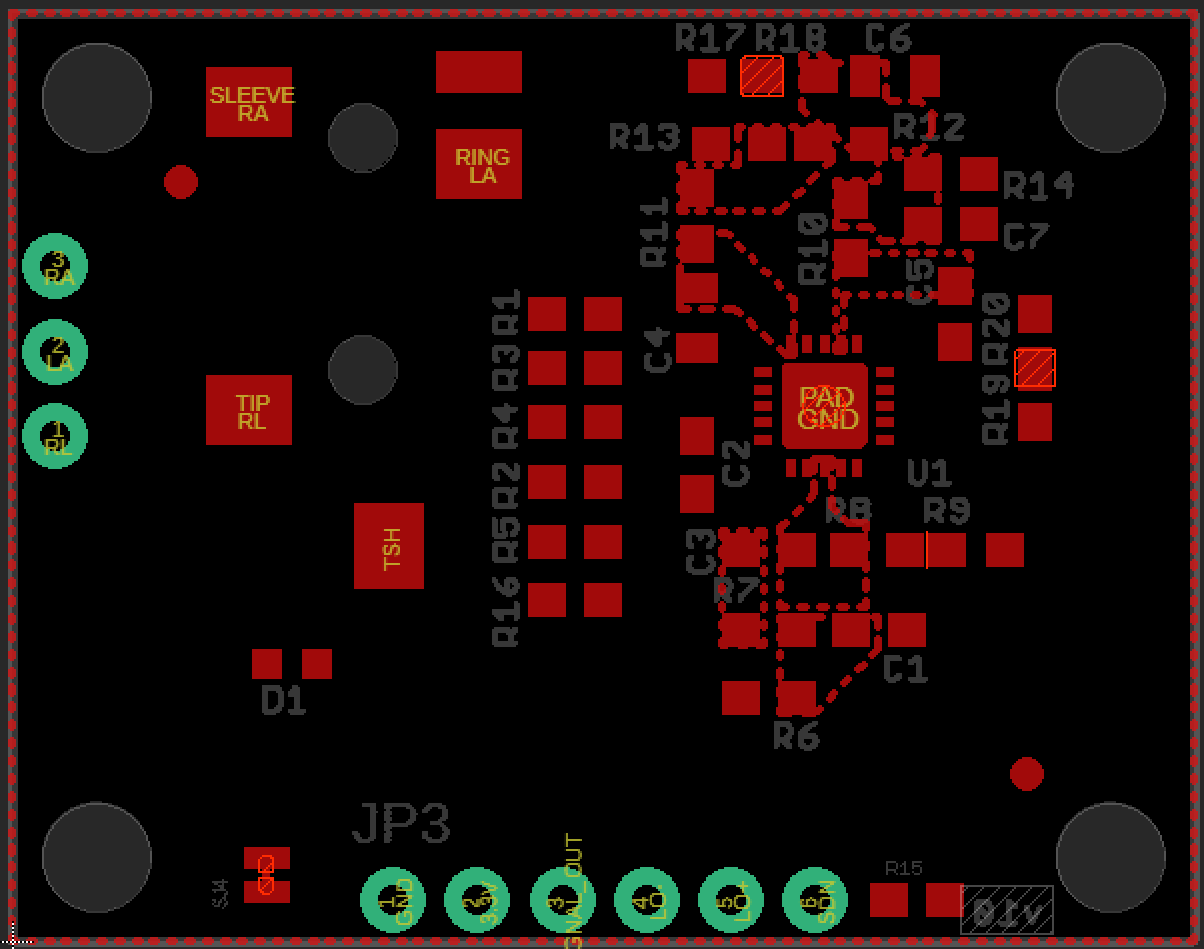}
    \subcaption{Layout Diagram by Industrial EDA Layout Engine}
  \end{minipage} \\[6pt]

  \begin{minipage}[t]{0.44\linewidth}
    \centering
    \includegraphics[width=\linewidth, height=9cm, keepaspectratio=false]{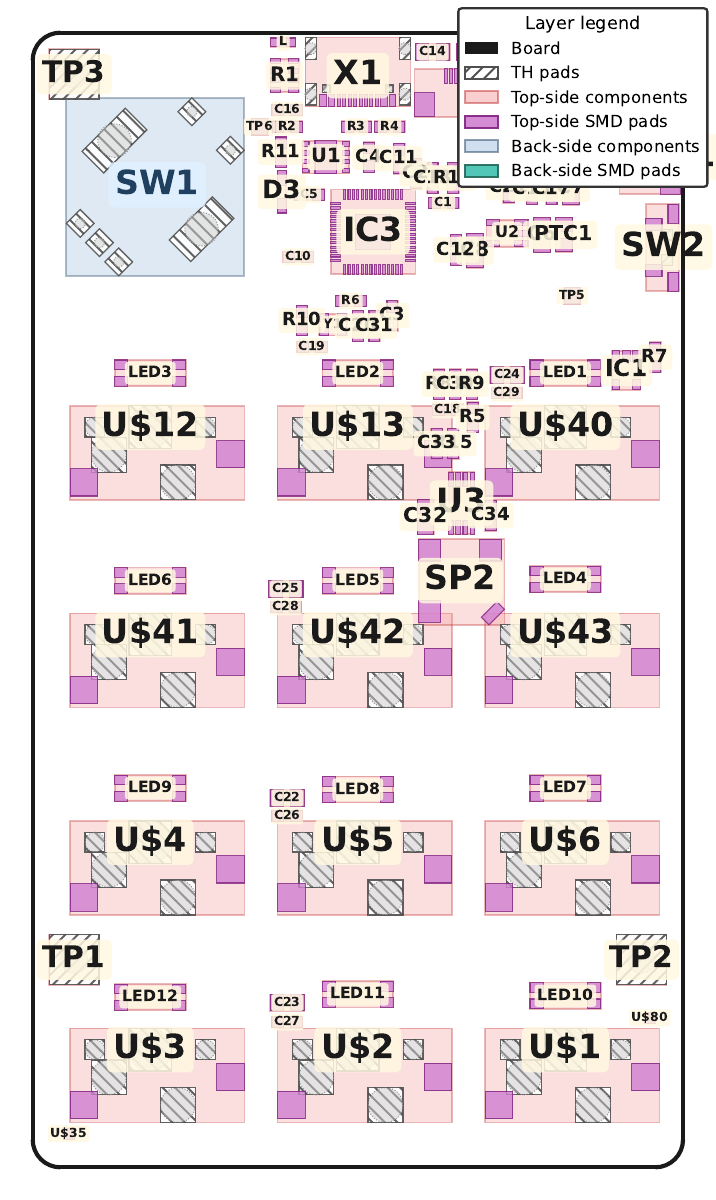}
    \subcaption{Layout Diagram by Custom EDA Layout Engine}
  \end{minipage} \hfill
  \begin{minipage}[t]{0.44\linewidth}
    \centering
    \includegraphics[width=\linewidth, height=9cm, keepaspectratio=false]{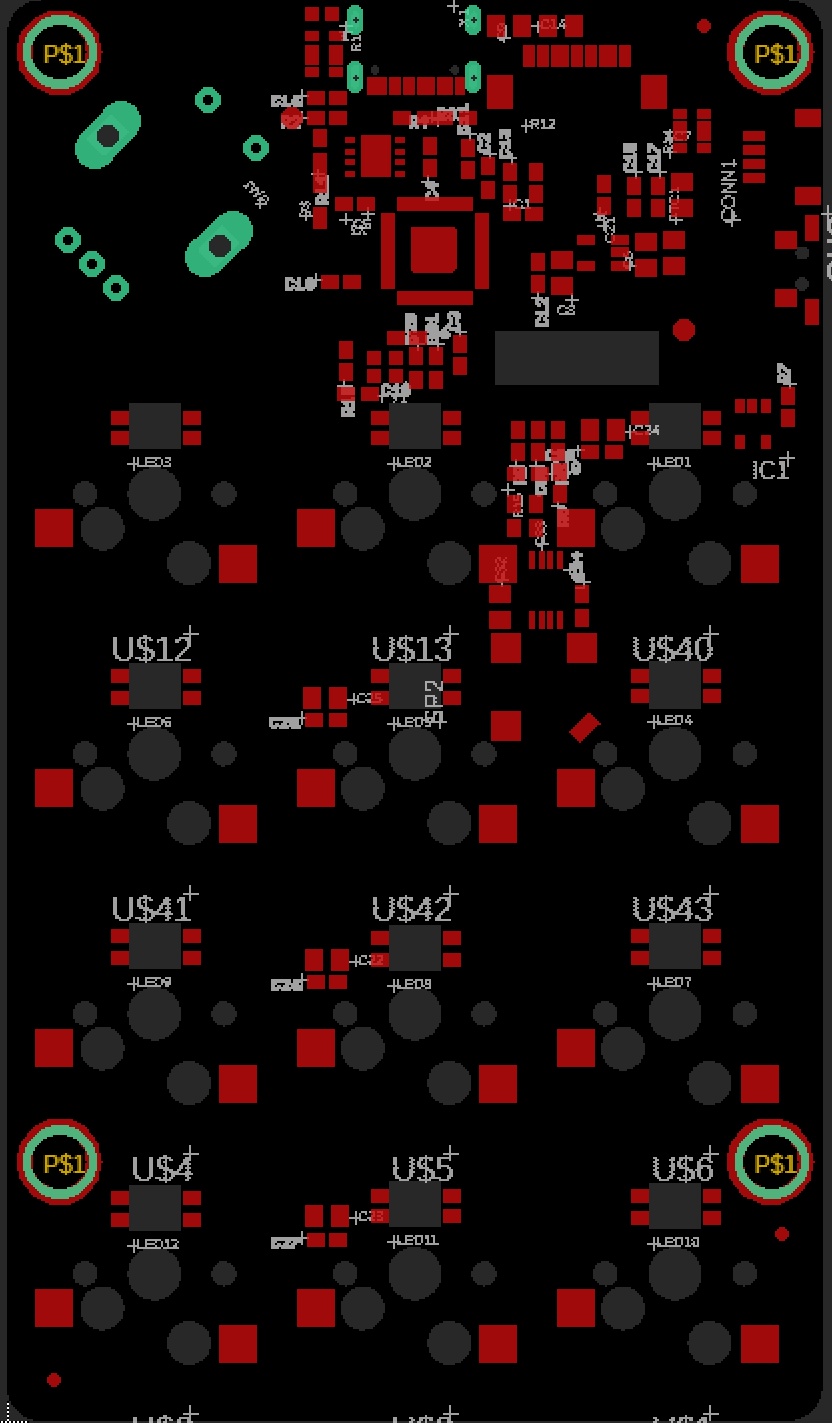}
    \subcaption{Layout Diagram by Industrial EDA Layout Engine}
  \end{minipage}

  \caption{Qualitative examples of our custom PCB layout rendering pipeline. Each row corresponds to one layout example. The left column shows layouts rendered by our custom EDA engine, and the right column shows the corresponding reference layouts produced by industrial EDA tools.}
  \label{fig:layout_examples}
\end{figure*}

\subsection{Automatic Annotation Pipeline}
\label{sec:appendix:pipeline}
To support scalable benchmark construction, we develop a source-driven annotation pipeline that extracts all required ground-truth information directly from EAGLE XML project files, covering both schematic-side and layout-side annotations without manual image-space labeling. Our pipeline operates on the XML-based format used by Autodesk EAGLE 9.6.2, which explicitly encodes component instances, symbol and footprint definitions, pin attributes, net topology, layer assignments, orientations, and board frame geometry as structured data.
\paragraph{Schematic Annotation.}
For schematic-side annotations, we adopt the rendering engine from OmniSch~\cite{lu2026omnisch}, which parses EAGLE XML schematic files to produce high-fidelity schematic diagrams paired with pixel-aligned annotations. Given a schematic project, the engine recovers the hierarchical circuit structure including symbol libraries, instance placements, pin attributes, and net topology, and programmatically reconstructs the schematic diagram by reproducing the native EDA visualization of symbols, wires, and text elements. Since each rendered visual primitive remains linked to its underlying semantic entity during image generation, the pipeline automatically extracts image-aligned annotations for symbol instances, pins, textual attributes, net labels, wires, and junctions, along with structural metadata required for downstream topology reasoning. These schematic annotations serve as reference signals for evaluating electrical functionality preservation in placement results.
\paragraph{Layout Annotation.}
For layout-side annotations, we develop a dedicated PCB layout rendering engine that parses ground-truth placement information directly from EAGLE XML layout files. Given a layout project, the engine first recovers the complete placement representation including component instance names, reference footprint definitions, pad-level block dimensions, net topology, and per-component position and rotation attributes. Each component instance is associated with its footprint definition, from which we extract block-level geometry including pad type (\texttt{smd}$/$\texttt{pad}$/$\texttt{hole}), physical dimensions, and layer assignment (\texttt{front}$/$\texttt{back}$/$\texttt{through}). Board frame geometry is separately recovered from DXF-encoded boundary primitives embedded in the layout file, from which we compute the bounding extents used to constrain all placement coordinates. The engine then programmatically reconstructs the PCB layout diagram, with each rendered element remaining linked to its underlying semantic entity, enabling automatic extraction of pixel-aligned annotations for component bounding boxes, pad locations, layer assignments, rotation codes, net connections, and board boundary primitives.
\paragraph{Unified Pipeline.}
Both annotation streams are processed within a unified pipeline that aligns schematic-side connectivity signals with layout-side placement ground truth through shared component instance identifiers. This cross-file correspondence enables automatic construction of paired training and evaluation samples covering spatial, routability, and electrical functionality dimensions without any manual intervention. Compared with purely manual image-space annotation, this source-driven strategy improves annotation consistency, eliminates geometric ambiguity arising from dense component layouts, and scales efficiently to large benchmark collections. Beyond static annotation extraction, the layout rendering engine additionally serves as an interactive visualization tool within our agentic evaluation framework, rendering the current placement state as visual feedback to support iterative placement refinement by large multimodal models.

\subsection{Structured Annotation Schema}
\label{sec:annotation_schema}

The annotations in \name are organized into three levels with
different annotation objects and counting units.

\paragraph{Entity-level annotations.}
This level describes individual objects in the PCB layout and
schematic. For each layout component, we provide its identity,
value, package, reference placement, orientation, layer assignment,
and bounding-box information. For each pad, we provide its identity,
type, dimensions, position in the component coordinate frame, layer,
and associated component. The corresponding schematic symbols and
pins are annotated with their identities, positions, textual
attributes, and component associations. Shared identifiers establish
explicit correspondences between layout components and pads and
their schematic symbols and pins. In total, the benchmark contains
\tai{77.24K} component placements and \tai{168.82K} pads.

\paragraph{Netlist-graph annotations.}
This level represents circuit topology. The benchmark contains
\tai{55.83K} netlist graphs whose nodes correspond to layout
components and pads or their schematic symbols and pins. Graph
connectivity records the electrical relationships among these
entities, including pin--pad correspondences, pad--pad connections,
and semantic net labels when available. These annotations provide
an unweighted structural representation of circuit connectivity.

\paragraph{Spatially weighted net annotations.}
This level augments the graph topology with spatial information.
Across the netlist graphs, the benchmark contains \tai{139.4K} nets.
Each net has two aligned spatial representations: a layout
representation computed from component and pad positions in the
expert-designed reference layout, and a schematic representation
computed from the positions of the corresponding symbols and pins
in the schematic coordinate frame. These representations provide
layout-based and schematic-based spatial weights for evaluating
reference-layout agreement and schematic-relationship preservation,
respectively.

The two dataset statistics therefore refer to different objects:
\tai{55.83K} is the number of netlist graphs, whereas
\tai{139.4K} is the number of nets contained in those graphs.
Spatially weighted annotations measure relational organization and
should not be interpreted as verification of electrical correctness.

\subsection{Visualization Example of Custom Layout Engine}
To qualitatively illustrate the capability of our custom EDA layout rendering engine, we provide representative examples in Figure~\ref{fig:layout_examples}. For each example, we show the corresponding industrial EDA layout rendered by Autodesk EAGLE alongside the synthesized layout rendered by our custom engine. This side-by-side comparison highlights the fidelity of our rendering pipeline in reproducing component placements, pad geometries, board boundaries, and routing traces. In particular, the examples demonstrate that our synthesized layouts faithfully preserve the structural and spatial characteristics of the original industrial designs, while additionally enabling dense instance-level annotation of component bounding boxes, pad locations, layer assignments, net connections, and board boundary primitives.

\section{Task Formulation Protocol}\label{sec:appendix:eval}
In this section we introduce the definition of each task, and evaluation metrics for each task. 

\subsection{Task Definition}
We formulate PCB component placement evaluation across the following metrics, each targeting a distinct aspect of layout quality.

\textbf{Overlap (OO)} refers to the normalized intersecting area between component pairs in the predicted placement, measuring the degree to which components physically conflict with one another on the board.

\textbf{Out-of-Boundary (OoB)} refers to the ratio of component area placed outside the board boundary $B$, capturing whether the predicted placement respects the spatial extent of the board.

\textbf{Half-Perimeter Wirelength (HPWL)} refers to the sum of bounding box half-perimeters over all nets, serving as a pre-routing proxy for the total wiring cost of the predicted placement. Lower HPWL indicates more compact and routing-friendly component arrangement.

\textbf{Net Crossing (NC)} refers to the total number of intersections between net connections in the predicted placement, where a lower count indicates fewer routing conflicts and a cleaner signal topology.

\textbf{Net Separation (NS)} refers to the average spatial distance between electrically connected pins across all nets, measuring how well the placement clusters connected components together.

\textbf{Predicted Routed Wirelength (PRWL)} refers to the actual total wirelength after routing via FreeRouting~\cite{freerouting2024}, capturing the true routing cost of the predicted placement.

\textbf{Routability Ratio (RR)} refers to the fraction of nets successfully routed by FreeRouting~\cite{freerouting2024}, directly measuring whether the predicted placement supports complete electrical connectivity.

\textbf{Number of Vias (\#Vias)} refers to the total number of vias introduced during routing, where a lower count indicates a more layer-efficient placement with fewer inter-layer transitions.

\textbf{Schematic Relational Distance ($\mathcal{D}_{sch}$)} refers to the average pairwise relative distance deviation between the predicted layout and the schematic diagram over all component pairs within the same subcircuit, measuring whether the predicted placement preserves the visual subcircuit topology of the circuit schematic.

\textbf{Reference Distance ($\mathcal{D}_{ref}$)} refers to the average pairwise relative distance deviation between the predicted layout and the expert-designed reference layout, measuring the degree to which the predicted placement aligns with professional component arrangement decisions.

\textbf{Constraint Satisfaction Rate (CSR)} refers to the ratio of evaluated geometric and routing constraints satisfied at the final placement iteration, measuring the overall correctness of the agentic placement outcome.

\textbf{Step Efficiency (SE)} refers to the average number of tool-use steps required to reach a valid placement, where a lower value indicates more efficient constraint-aware spatial reasoning.

\textbf{Placement Refinement Quality (PRQ)} refers to the fraction of iterations in which the placement improves over the previous step, capturing whether the agent consistently makes progress toward a valid placement across iterations.

\subsection{Evaluation Metrics}

Once the component placement is predicted, we compute the following evaluation metrics across four complementary dimensions.

\textbf{Overlap (OO).} Given a predicted placement $\mathcal{P}$ with components $\{c_1, c_2, \ldots, c_N\}$, the normalized overlapping area between components is computed as:
\begin{equation}
\text{OO}(\mathcal{P}) = \frac{\sum_{i < j} \text{Area}(c_i \cap c_j)}{\sum_{i} \text{Area}(c_i)}
\end{equation}
where $c_i \cap c_j$ denotes the intersecting area between components $i$ and $j$. Lower values indicate fewer physical conflicts between components. We only compute for components in the same layer.

\textbf{Out-of-Boundary (OoB).} The ratio of component area placed outside the board boundary $B$ is computed as:
\begin{equation}
\text{OoB}(\mathcal{P}) = \frac{\sum_{i} \text{Area}(c_i \setminus B)}{\sum_{i} \text{Area}(c_i)}
\end{equation}
where $c_i \setminus B$ denotes the area of component $i$ outside the board boundary. Lower values indicate better spatial containment within the board.

\textbf{Half-Perimeter Wirelength (HPWL).} For a set of nets $\mathcal{N}$, HPWL is computed as:
\begin{equation}
\begin{split}
\text{HPWL}(\mathcal{P}) = \sum_{n \in \mathcal{N}} & \left( \max_{i \in n} x_i - \min_{i \in n} x_i \right. \\
& \left. + \max_{i \in n} y_i - \min_{i \in n} y_i \right)
\end{split}
\end{equation}
where $(x_i, y_i)$ denotes the position of pin $i$ in net $n$. Lower HPWL indicates more compact and routing-friendly component placement.

\textbf{Net Crossing (NC)} refers to the total number of intersections between net connections in the predicted placement, where a lower count indicates fewer routing conflicts and a cleaner signal topology:
\begin{equation}
\text{NC}(\mathcal{P}) = \sum_{n_i \in \mathcal{N}} \sum_{n_j \in \mathcal{N}, j > i} \mathbb{1}\left[ e_{n_i} \cap e_{n_j} \neq \emptyset \right]
\end{equation}
where $e_{n_i}$ and $e_{n_j}$ denote the straight-line connections of nets $n_i$ and $n_j$ respectively, and $\mathbb{1}[\cdot]$ is the indicator function that equals 1 when two net connections intersect.

\textbf{Net Separation (NS).} NS measures the average spatial distance between electrically connected pins across all nets:
\begin{equation}
\text{NS}(\mathcal{P}) = \frac{1}{|\mathcal{N}|}\sum_{n \in \mathcal{N}} \frac{1}{|n|} \sum_{i \in n} \| p_i - \bar{p}_n \|_2
\end{equation}
where $p_i$ denotes the position of pin $i$ and $\bar{p}_n$ denotes the centroid of net $n$.

\textbf{Routability Ratio (RR).} RR is defined as the fraction of successfully routed nets obtained via FreeRouting~\cite{freerouting2024}:
\begin{equation}
\text{RR}(\mathcal{P}) = \frac{|\mathcal{N}_{routed}|}{|\mathcal{N}|}
\end{equation}
where $|\mathcal{N}_{routed}|$ denotes the number of successfully routed nets and $|\mathcal{N}|$ denotes the total number of nets.

\textbf{Predicted Routed Wirelength (PRWL)} refers to the actual total wirelength after routing via FreeRouting~\cite{freerouting2024}, capturing the true routing cost of the predicted placement:
\begin{equation}
\text{PRWL}(\mathcal{P}) = \sum_{n \in \mathcal{N}} \sum_{e \in \mathcal{R}_n} \|e\|_2
\end{equation}
where $\mathcal{R}_n$ denotes the set of routed wire segments for net $n$, and $\|e\|_2$ denotes the Euclidean length of wire segment $e$.

\textbf{Number of Vias (\#Vias).} \#Vias records the total number of vias introduced during routing, where a lower count indicates a more layer-efficient placement.

\textbf{Schematic Relational Distance ($\mathcal{D}_{sch}$).} 
Since PCB component placement does not admit a unique ground truth solution, we measure the visual relative distance between the predicted layout and the schematic diagram $\mathcal{S}$:
\begin{equation}
\mathcal{D}_{sch}(\mathcal{P}, \mathcal{S}) = \frac{1}{|\mathcal{G}|}\sum_{(i,j) \in \mathcal{G}} \left| d_{\mathcal{P}}(i,j) - d_{\mathcal{S}}(i,j) \right|
\end{equation}
where $\mathcal{G}$ denotes the set of component pairs within the same subcircuit, and $d_{\mathcal{P}}(i,j)$, $d_{\mathcal{S}}(i,j)$ denote the pairwise distances between components $i$ and $j$ in the predicted layout and schematic diagram respectively.

\textbf{Reference Distance ($\mathcal{D}_{ref}$).} We further measure the component-level relative placement distance against the expert-designed reference layout $\mathcal{R}$:
\begin{equation}
\mathcal{D}_{ref}(\mathcal{P}, \mathcal{R}) = \frac{1}{|\mathcal{G}|}\sum_{(i,j) \in \mathcal{G}} \left| d_{\mathcal{P}}(i,j) - d_{\mathcal{R}}(i,j) \right|
\end{equation}
where $d_{\mathcal{R}}(i,j)$ denotes the pairwise distance between components $i$ and $j$ in the reference layout. Because lower $D_{\mathrm{ref}}$ indicates closer agreement, we report
the bounded reference-layout similarity
\begin{equation}
S_{\mathrm{ref}}(\mathcal{P},\mathcal{R})
=
\frac{1}{1+D_{\mathrm{ref}}(\mathcal{P},\mathcal{R})}.
\end{equation}
Higher $S_{\mathrm{ref}}$ indicates closer agreement with the expert
layout. This metric is an auxiliary spatial diagnostic and does not
measure electrical correctness.

\textbf{Constraint Satisfaction Rate (CSR).} CSR measures the ratio of evaluated geometric and routing constraints satisfied at the final placement iteration $T$:
\begin{equation}
\text{CSR}(\mathcal{P}_T) = \frac{|\mathcal{C}_{sat}|}{|\mathcal{C}|}
\end{equation}
where $|\mathcal{C}_{sat}|$ and $|\mathcal{C}|$ denote the number of satisfied and total constraints respectively.

\textbf{Placement Refinement Quality (PRQ).} PRQ measures whether successive placement iterations progressively reduce constraint violations:
\begin{equation}
\text{PRQ} = \frac{1}{T-1}\sum_{t=1}^{T-1} \mathbb{1}\left[ \mathcal{V}(\mathcal{P}_{t+1}) < \mathcal{V}(\mathcal{P}_t) \right]
\end{equation}
where $\mathcal{V}(\mathcal{P}_t)$ denotes the total number of constraint violations at iteration $t$.

\textbf{Functional-Grouping AUC.}
For each board, component pairs are labeled positive when they belong
to the same schematic functional group and negative otherwise. We rank
pairs using negative Euclidean placement distance and compute ROC AUC.
Higher AUC indicates that same-group components tend to be placed
closer together. This metric allows multiple coordinate solutions but
remains a relational proxy rather than electrical validation.

\textbf{Step Efficiency (SE).} SE measures the average number of tool-use steps required to reach a valid placement, where a lower value indicates more efficient constraint-aware reasoning.

\section{Experiment}
\label{sec:appendix:exp}

\subsection{Additional Analysis of Agentic Tool Use}
\label{sec:agentic_analysis}

The agentic study evaluates three complementary feedback channels.
Visualization provides a rendered image of the current placement;
Geo. Check provides explicit information about component-overlap and
boundary violations; and Routing Score provides HPWL-based routing
feedback. The All Tools condition makes all three feedback channels
available simultaneously.

The results in Table~\ref{tab:agentic} demonstrate that feedback
utility depends on both the model and the optimization objective. For
the three models examined here, the All Tools condition achieves the
lowest overlap error. Claude-Sonnet-4.6 obtains an overlap error of
$0.004$ with All Tools, compared with $0.029$ using Visualization,
$0.011$ using Geo. Check, and $0.055$ using Routing Score.
Ministral-14B similarly achieves its lowest overlap error with All
Tools ($0.065$), compared with $0.130$, $0.177$, and $0.124$ under
Visualization, Geo. Check, and Routing Score, respectively.
GPT-5-mini follows the same pattern, achieving an overlap error of
$0.024$ with All Tools, compared with $0.028$ using Visualization,
$0.049$ using Geo. Check, and $0.095$ using Routing Score. These
results suggest that jointly providing visual, geometric, and routing
feedback can facilitate overlap correction by allowing models to
identify and repair layout conflicts using complementary signals.

However, the All Tools condition does not consistently achieve the
best performance across other evaluation metrics. For example,
Visualization produces the lowest out-of-boundary ratio for
Claude-Sonnet-4.6 and Ministral-14B, whereas Routing Score produces
the lowest HPWL and highest routability ratio for Ministral-14B.
Similarly, although All Tools gives GPT-5-mini its lowest overlap error
and HPWL, Visualization yields a substantially lower out-of-boundary
ratio. These results reveal trade-offs among geometric legality,
routability, and relational placement quality: optimizing one
criterion may degrade another.

Overall, combining all feedback channels is effective for reducing
overlap for the models analyzed here, but it is not uniformly optimal
across models and evaluation objectives. Simultaneous feedback may
introduce competing objectives or cause a model to prioritize one type
of violation over another. These findings motivate adaptive agentic
systems that select feedback channels according to the current layout
state, target metric, and model capability. Because the successfully
parsed sample subsets may differ across conditions, we interpret these
results as within-agentic diagnostic comparisons rather than evidence
of uniform improvement over one-shot placement.

\subsection{Prompt Template}
We use task-specific prompt templates to guide the model under different PCB layout settings. Rather than using a single generic instruction, each task is associated with a tailored prompt that consolidates the placement objective, board boundary constraints, component block dimensions, netlist connectivity, allowed rotation codes, placement rules, and required output format into a single unified instruction. The model is expected to parse all spatial and structural information within this self-contained prompt and produce a valid, constraint-satisfying placement in a single pass. In the following, we present representative prompt examples used in our layout evaluation framework.

\noindent\textbf{Oneshot Evaluation Prompt Template.}
We design five prompt variants to systematically evaluate the impact of different auxiliary inputs on LLM-based PCB placement. $\bullet$ \textbf{Multimodal (Schematic Image-Guided)} This variant extends the baseline by supplying rendered schematic images as additional visual input, allowing the model to leverage visual circuit structure as a reference for functional grouping and relative placement hints.
$\bullet$ \textbf{Graph-Augmented (Schematic Graph-Guided)} This variant injects a secondary schematic graph JSON encoding symbol-level connectivity with spatial distance weights on pin-pin edges, guiding the model to keep strongly-related components closer together on the board.
$\bullet$ \textbf{Full-Graph (Spatial-Semantic Graph-Guided)} This variant further enriches the graph representation by retaining per-node semantic attributes for symbols, pins, nets, and text annotations alongside the spatial edge weights, enabling the model to jointly exploit both structural connectivity and semantic context.
$\bullet$ \textbf{Few-Shot (Example-Guided)} This variant augments the baseline with one concrete placement example demonstrating an effective layout strategy, instructing the model to mimic the observed spatial reasoning patterns while adapting coordinates to the current board and component set.

\begin{tcolorbox}[
    enhanced,
    breakable,
    colback=casebluebg,
    colframe=caseblueborder,
    boxrule=0.6pt,
    arc=0pt,
    outer arc=0pt,
    left=10pt,
    right=10pt,
    top=8pt,
    bottom=8pt,
]
\ttfamily\small
\#\textbf{Oneshot\_prompt (Baseline Task):}\\
You are an expert PCB layout engineer. Your task is to place all components\\
on a PCB board optimally.\\
\\
\textbf{Board Boundary:}\\
Overall Board boundary extents: ($x_0$, $y_0$) to ($x_1$, $y_1$), size $W$ mm $\times$ $H$ mm.\\
Board outline is defined only by \texttt{board\_boundary.dxf\_details}.\\
Use this outline to constrain all placements inside the board.\\
Board boundary \texttt{dxf\_details} (JSON): \{dxf\_details\_json\}\\
\\
\textbf{Component Block Dimensions:}\\
For each component, \texttt{block\_dimension} is a list of blocks inside the
component that cannot overlap with blocks from other components.
Each block has: position (relative to component center), width, length,
type (\texttt{smd}/\texttt{pad}/\texttt{hole}), and layer (\texttt{front}/\texttt{back}/\texttt{through}).\\
Allowed rotation codes: 
\quad \texttt{R0}: front side, 0\textdegree \quad
\texttt{R180}: front side, 180\textdegree \quad
\texttt{R270}: front side, 270\textdegree\\
\quad \texttt{MR0}: back side (mirrored), 0\textdegree \quad
\texttt{MR180}: back side (mirrored), 180\textdegree \quad
\texttt{MR270}: back side (mirrored), 270\textdegree\\
\\
Components (name and block dimensions):\\
\quad - \textit{COMP\_NAME}: \textit{type} (\textit{layer}) $W$\,mm $\times$ $L$\,mm at ($x$, $y$)\\
\quad - $\cdots$\\
\\
\textbf{Netlist (connections):}\\
\quad - Net `\textit{NET\_NAME}': \textit{element}:\textit{pad}, \textit{element}:\textit{pad}, $\ldots$\\
\quad - $\cdots$\\
\\
\textbf{Placement Rules:}\\
1. Coordinates must be inside the board boundary\\
\quad ($x_0 \leq x \leq x_1$,\; $y_0 \leq y \leq y_1$).\\
2. \textbf{CRITICAL}: Block dimensions from different components must NOT overlap.
\quad Blocks within the same component are already non-overlapping by design.\\
3. Components CAN share the same center position, but their\\
\quad \texttt{block\_dimensions} must not overlap in 2D.\\
4. Choose placement side (\texttt{R*} vs \texttt{MR*}) to avoid block collisions
\quad while respecting layer constraints.\\
\\
\textbf{Output Format} (strictly follow this):\\
Return ONLY a valid JSON object, no explanation, no markdown fences.\\
\{\\
\quad ``placements'': [\\
\quad\quad \{``name'': ``COMP'', ``x'': \textit{X}, ``y'': \textit{Y}, ``rotation'': ``R0$|\cdots$''\},\\
\quad\quad $\cdots$\\
\quad ]\\
\}\\
Every component must appear exactly once. Coordinates are in millimeters.\\
Place component CENTER at ($x$,\,$y$).
\end{tcolorbox}

\begin{tcolorbox}[
    enhanced,
    breakable,
    colback=casegreenbg,
    colframe=casegreenborder,
    boxrule=0.6pt,
    arc=0pt,
    outer arc=0pt,
    left=10pt,
    right=10pt,
    top=8pt,
    bottom=8pt,
]
\ttfamily\small
\#\textbf{Multimodal\_prompt (Schematic Image-Guided Layout Task):}\\
You are an expert PCB layout engineer. Your task is to place all components\\
on a PCB board optimally.\\
\\
You are also given schematic image(s) as multimodal input in this request.\\
Treat them as reference for functional grouping and relative placement hints.\\
\\
\textbf{Board Boundary:}\\
Overall Board boundary extents: ($x_0$, $y_0$) to ($x_1$, $y_1$), size $W$ mm $\times$ $H$ mm.
Board outline is defined only by \texttt{board\_boundary.dxf\_details}.
Use this outline to constrain all placements inside the board.
Board boundary \texttt{dxf\_details} (JSON): \{dxf\_details\_json\}\\
\\
\textbf{Component Block Dimensions:}\\
For each component, \texttt{block\_dimension} is a list of blocks inside the
component that cannot overlap with blocks from other components.
Each block has: position (relative to component center), width, length,
type (\texttt{smd}/\texttt{pad}/\texttt{hole}), and layer (\texttt{front}/\texttt{back}/\texttt{through}).\\
Allowed rotation codes:
\quad \texttt{R0}: front side, 0\textdegree \quad
\texttt{R180}: front side, 180\textdegree \quad
\texttt{R270}: front side, 270\textdegree \\
\quad \texttt{MR0}: back side (mirrored), 0\textdegree \quad
\texttt{MR180}: back side (mirrored), 180\textdegree \quad
\texttt{MR270}: back side (mirrored), 270\textdegree
\\
Components (name and block dimensions):\\
\quad - \textit{COMP\_NAME}: \textit{type} (\textit{layer}) $W$\,mm $\times$ $L$\,mm at ($x$, $y$)\\
\quad - $\cdots$\\
\\
\textbf{Netlist (connections):}\\
\quad - Net `\textit{NET\_NAME}': \textit{element}:\textit{pad}, \textit{element}:\textit{pad}, $\ldots$\\
\quad - $\cdots$\\
\\
\textbf{Placement Rules:}\\
1. Coordinates must be inside the board boundary\\
\quad ($x_0 \leq x \leq x_1$,\; $y_0 \leq y \leq y_1$).\\
2. \textbf{CRITICAL}: Block dimensions from different components must NOT overlap.
\quad Blocks within the same component are already non-overlapping by design.\\
3. Components CAN share the same center position, but their
\quad \texttt{block\_dimensions} must not overlap in 2D.\\
4. Choose placement side (\texttt{R*} vs \texttt{MR*}) to avoid block collisions\\
\quad while respecting layer constraints.\\
5. Use the provided schematic image(s) as reference for functional grouping
\quad and relative placement hints between connected components.\\
\\
\textbf{Output Format} (strictly follow this):\\
Return ONLY a valid JSON object, no explanation, no markdown fences.\\
\{\\
\quad ``placements'': [\\
\quad\quad \{``name'': ``COMP'', ``x'': \textit{X}, ``y'': \textit{Y}, ``rotation'': ``R0$|\cdots$''\},\\
\quad\quad $\cdots$\\
\quad ]\\
\}\\
Every component must appear exactly once. Coordinates are in millimeters.\\
Place component CENTER at ($x$,\,$y$).
\end{tcolorbox}

\definecolor{caseorangebg}{RGB}{255, 248, 235}
\definecolor{caseorangeborder}{RGB}{210, 130, 40}
\definecolor{casegreenbg}{RGB}{235, 255, 235}

\begin{tcolorbox}[
    enhanced,
    breakable,
    colback=caseorangebg,
    colframe=caseorangeborder,
    boxrule=0.6pt,
    arc=0pt,
    outer arc=0pt,
    left=10pt,
    right=10pt,
    top=8pt,
    bottom=8pt,
]
\ttfamily\small
\#\textbf{Graph-Augmented\_prompt (Schematic Graph-Guided Layout Task):}\\
You are an expert PCB layout engineer. Your task is to place all components
on a PCB board optimally. \\

You are also provided with a schematic
graph (JSON) extracted from the schematic file, which encodes symbol-level
connectivity and relative spatial relationships among components. Use it as
an auxiliary signal to guide functional grouping and placement decisions.\\
\\
\textbf{Board Boundary:}\\
Overall Board boundary extents: ($x_0$, $y_0$) to ($x_1$, $y_1$), size $W$ mm $\times$ $H$ mm.
Board outline is defined only by \texttt{board\_boundary.dxf\_details}.
Use this outline to constrain all placements inside the board.
Board boundary \texttt{dxf\_details} (JSON): \{dxf\_details\_json\}\\
\\
\textbf{Component Block Dimensions:}\\
For each component, \texttt{block\_dimension} is a list of blocks inside the
component that cannot overlap with blocks from other components.
Each block has: position (relative to component center), width, length,
type (\texttt{smd}/\texttt{pad}/\texttt{hole}), and layer (\texttt{front}/\texttt{back}/\texttt{through}).\\
Allowed rotation codes:\\
\quad \texttt{R0}: front side, 0\textdegree \quad
\texttt{R180}: front side, 180\textdegree \quad
\texttt{R270}: front side, 270\textdegree\\
\quad \texttt{MR0}: back side (mirrored), 0\textdegree \quad
\texttt{MR180}: back side (mirrored), 180\textdegree \quad
\texttt{MR270}: back side (mirrored), 270\textdegree\\
\\
Components (name and block dimensions):\\
\quad - \textit{COMP\_NAME}: \textit{type} (\textit{layer}) $W$\,mm $\times$ $L$\,mm at ($x$, $y$)\\
\quad - $\cdots$\\
\\
\textbf{Netlist (connections):}\\
\quad - Net `\textit{NET\_NAME}': \textit{element}:\textit{pad}, \textit{element}:\textit{pad}, $\ldots$\\
\quad - $\cdots$\\
\\
\textbf{Secondary Schematic Graph Input (JSON):}\\
\{sch\_graph\_json\}\\
\\
How to interpret this schematic graph:\\
\quad - Netlist information extracted from the schematic (\texttt{.sch}) file.\\
\quad - Edges with type \texttt{symbol\_pin} indicate which symbol each pin belongs to.\\
\quad - Edges with type \texttt{pin-pin} indicate connectivity among symbols\\
\quad\quad through their pins.\\
\quad - For \texttt{pin-pin} edges, the weight is a relative distance signal on\\
\quad\quad the schematic image. Smaller weight means symbols are relatively\\
\quad\quad closer on the schematic.\\
\quad - Use this relative-distance signal to improve placement decisions:\\
\quad\quad keep strongly-related symbols relatively close when possible.\\
\\
\textbf{Placement Rules:}\\
1. Coordinates must be inside the board boundary\\
\quad ($x_0 \leq x \leq x_1$,\; $y_0 \leq y \leq y_1$).\\
2. \textbf{CRITICAL}: Block dimensions from different components must NOT overlap.\\
\quad Blocks within the same component are already non-overlapping by design.\\
3. Components CAN share the same center position, but their\\
\quad \texttt{block\_dimensions} must not overlap in 2D.\\
4. Choose placement side (\texttt{R*} vs \texttt{MR*}) to avoid block collisions\\
\quad while respecting layer constraints.\\
\\
\textbf{Output Format} (strictly follow this):\\
Return ONLY a valid JSON object, no explanation, no markdown fences.\\
\{\\
\quad ``placements'': [\\
\quad\quad \{``name'': ``COMP'', ``x'': \textit{X}, ``y'': \textit{Y}, ``rotation'': ``R0$|\cdots$''\},\\
\quad\quad $\cdots$\\
\quad ]\\
\}\\
Every component must appear exactly once. Coordinates are in millimeters.\\
Place component CENTER at ($x$,\,$y$).
\end{tcolorbox}

\definecolor{casetealbg}{RGB}{235, 248, 250}
\definecolor{casetealborder}{RGB}{40, 150, 160}

\begin{tcolorbox}[
    enhanced,
    breakable,
    colback=casetealbg,
    colframe=casetealborder,
    boxrule=0.6pt,
    arc=0pt,
    outer arc=0pt,
    left=10pt,
    right=10pt,
    top=8pt,
    bottom=8pt,
]
\ttfamily\small
\#\textbf{Full-Graph\_prompt (Spatial-Semantic Graph-Guided Layout Task):}\\
You are an expert PCB layout engineer. Your task is to place all components\\
on a PCB board optimally.\\
You are also provided with a secondary schematic graph (JSON) extracted from\\
the schematic file, which jointly encodes symbol-level connectivity with\\
relative spatial distance weights and per-node semantic attributes for\\
symbols, pins, nets, and text annotations. Use both signals together as\\
auxiliary guidance for functional grouping and placement decisions.\\
\\
\textbf{Board Boundary:}\\
Overall Board boundary extents: ($x_0$, $y_0$) to ($x_1$, $y_1$), size $W$ mm $\times$ $H$ mm.
Board outline is defined only by \texttt{board\_boundary.dxf\_details}.
Use this outline to constrain all placements inside the board.
Board boundary \texttt{dxf\_details} (JSON): \{dxf\_details\_json\}\\
\\
\textbf{Component Block Dimensions:}\\
For each component, \texttt{block\_dimension} is a list of blocks inside the
component that cannot overlap with blocks from other components.
Each block has: position (relative to component center), width, length,
type (\texttt{smd}/\texttt{pad}/\texttt{hole}), and layer (\texttt{front}/\texttt{back}/\texttt{through}).\\
Allowed rotation codes:\\
\quad \texttt{R0}: front side, 0\textdegree \quad
\texttt{R180}: front side, 180\textdegree \quad
\texttt{R270}: front side, 270\textdegree\\
\quad \texttt{MR0}: back side (mirrored), 0\textdegree \quad
\texttt{MR180}: back side (mirrored), 180\textdegree \quad
\texttt{MR270}: back side (mirrored), 270\textdegree\\
\\
Components (name and block dimensions):\\
\quad - \textit{COMP\_NAME}: \textit{type} (\textit{layer}) $W$\,mm $\times$ $L$\,mm at ($x$, $y$)\\
\quad - $\cdots$\\
\\
\textbf{Netlist (connections):}\\
\quad - Net `\textit{NET\_NAME}': \textit{element}:\textit{pad}, \textit{element}:\textit{pad}, $\ldots$\\
\quad - $\cdots$\\
\\
\textbf{Secondary Schematic Graph Input (JSON):}\\
\{sch\_graph\_json\}\\
\\
How to interpret this schematic graph:\\
\quad - Extracted from the schematic (\texttt{.sch}) file; contains both structural\\
\quad\quad edges and per-node semantic attributes.\\
\quad - Edges with type \texttt{symbol\_pin} indicate which symbol each pin belongs to.\\
\quad - Edges with type \texttt{pin-pin} indicate connectivity among symbols\\
\quad\quad through their pins.\\
\quad - For \texttt{pin-pin} edges, the weight is a relative distance signal on
\quad\quad the schematic image. Smaller weight means symbols are relatively\\
\quad\quad closer on the schematic.\\
\quad - Use this relative-distance signal to improve placement decisions:\\
\quad\quad keep strongly-related symbols relatively close when possible.\\
\quad - Node attributes (for symbols/pins/nets) and text annotations provide\\
\quad\quad semantic context; use them as auxiliary hints for functional grouping.\\
\\
\textbf{Placement Rules:}\\
1. Coordinates must be inside the board boundary\\
\quad ($x_0 \leq x \leq x_1$,\; $y_0 \leq y \leq y_1$).\\
2. \textbf{CRITICAL}: Block dimensions from different components must NOT overlap.\\
\quad Blocks within the same component are already non-overlapping by design.\\
3. Components CAN share the same center position, but their\\
\quad \texttt{block\_dimensions} must not overlap in 2D.\\
4. Choose placement side (\texttt{R*} vs \texttt{MR*}) to avoid block collisions\\
\quad while respecting layer constraints.\\
\\
\textbf{Output Format} (strictly follow this):\\
Return ONLY a valid JSON object, no explanation, no markdown fences.\\
\{\\
\quad ``placements'': [\\
\quad\quad \{``name'': ``COMP'', ``x'': \textit{X}, ``y'': \textit{Y}, ``rotation'': ``R0$|\cdots$''\},\\
\quad\quad $\cdots$\\
\quad ]\\
\}\\
Every component must appear exactly once. Coordinates are in millimeters.\\
Place component CENTER at ($x$,\,$y$).
\end{tcolorbox}

\definecolor{caseredbg}{RGB}{255, 242, 242}
\definecolor{caseredborder}{RGB}{180, 60, 60}

\begin{tcolorbox}[
    enhanced,
    breakable,
    colback=caseredbg,
    colframe=caseredborder,
    boxrule=0.6pt,
    arc=0pt,
    outer arc=0pt,
    left=10pt,
    right=10pt,
    top=8pt,
    bottom=8pt,
]
\ttfamily\small
\#\textbf{Few-Shot\_prompt (Example-Guided Layout Task):}\\
You are an expert PCB layout engineer. Your task is to place all components
on a PCB board optimally. \\

You are also provided with three placement
examples that demonstrate effective layout strategies. Mimic the placement
reasoning and spatial patterns shown in the examples, while adapting exact
coordinates to the current board and component set.\\
\\
\textbf{Board Boundary:}\\
Overall Board boundary extents: ($x_0$, $y_0$) to ($x_1$, $y_1$), size $W$ mm $\times$ $H$ mm.\\
Board outline is defined only by \texttt{board\_boundary.dxf\_details}.\\
Use this outline to constrain all placements inside the board.\\
Board boundary \texttt{dxf\_details} (JSON): \{dxf\_details\_json\}\\
\\
\textbf{Component Block Dimensions:}\\
For each component, \texttt{block\_dimension} is a list of blocks inside the\\
component that cannot overlap with blocks from other components.\\
Each block has: position (relative to component center), width, length,\\
type (\texttt{smd}/\texttt{pad}/\texttt{hole}), and layer (\texttt{front}/\texttt{back}/\texttt{through}).\\
Allowed rotation codes:\\
\quad \texttt{R0}: front side, 0\textdegree \quad
\texttt{R180}: front side, 180\textdegree \quad
\texttt{R270}: front side, 270\textdegree\\
\quad \texttt{MR0}: back side (mirrored), 0\textdegree \quad
\texttt{MR180}: back side (mirrored), 180\textdegree \quad
\texttt{MR270}: back side (mirrored), 270\textdegree\\
\\
Components (name and block dimensions):\\
\quad - \textit{COMP\_NAME}: \textit{type} (\textit{layer}) $W$\,mm $\times$ $L$\,mm at ($x$, $y$)\\
\quad - $\cdots$\\
\\
\textbf{Netlist (connections):}\\
\quad - Net `\textit{NET\_NAME}': \textit{element}:\textit{pad}, \textit{element}:\textit{pad}, $\ldots$\\
\quad - $\cdots$\\
\\
\textbf{Placement Rules:}\\
1. Coordinates must be inside the board boundary\\
\quad ($x_0 \leq x \leq x_1$,\; $y_0 \leq y \leq y_1$).\\
2. \textbf{CRITICAL}: Block dimensions from different components must NOT overlap.\\
\quad Blocks within the same component are already non-overlapping by design.\\
3. Components CAN share the same center position, but their\\
\quad \texttt{block\_dimensions} must not overlap in 2D.\\
4. Choose placement side (\texttt{R*} vs \texttt{MR*}) to avoid block collisions\\
\quad while respecting layer constraints.\\
\\
\textbf{Few-Shot Examples} (mimic strategy, adapt coordinates to current board):\\
Example 1: \textit{Two ICs with nearby decoupling capacitors}\\
\quad Input: Board 40$\times$30\,mm. Components: U1(MCU), U2(SENSOR), C1, C2, J1.\\
\quad\quad Nets: U1:VCC--C1:1, U2:VCC--C2:1, U1:SCL--U2:SCL, U1:IO--J1:1.\\
\quad Output: \{``placements'': [\{U1 at (16,16,R0)\}, \{U2 at (24,16,R0)\},\\
\quad\quad \{C1 at (16,12.5,R0)\}, \{C2 at (24,12.5,R0)\}, \{J1 at (35,15,R180)\}]\}\\
\quad $\cdots$\\
\\
\textbf{Output Format} (strictly follow this):\\
Return ONLY a valid JSON object, no explanation, no markdown fences.\\
\{\\
\quad ``placements'': [\\
\quad\quad \{``name'': ``COMP'', ``x'': \textit{X}, ``y'': \textit{Y}, ``rotation'': ``R0$|\cdots$''\},\\
\quad\quad $\cdots$\\
\quad ]\\
\}\\
Every component must appear exactly once. Coordinates are in millimeters.\\
Place component CENTER at ($x$,\,$y$).
\end{tcolorbox}

\noindent\textbf{Agentic Evaluation Prompt Template.}
We design four prompt variants to systematically evaluate the impact of different feedback signals on LLM-based agentic PCB placement. $\bullet$ \textbf{HPWL-Guided (Routability Risk Feedback)} This variant replaces the raw half-perimeter wirelength metric with a learned routability risk score as the iterative feedback signal, allowing the agent to optimize toward congestion-aware placement objectives rather than purely geometric wire minimization. $\bullet$ \textbf{Overlap-Guided (Constraint Violation Feedback)} This variant exposes only the overlap inspection tool during each iteration, providing the agent with binary constraint violation signals that directly indicate whether placed components physically intersect, guiding the model to prioritize feasibility over optimality in its revision decisions. $\bullet$ \textbf{Boundary-Guided (Out-of-Boundary Feedback)} This variant restricts feedback to out-of-boundary detection results, supplying the agent with spatial legality signals that indicate whether components exceed board extents, encouraging the model to learn boundary-respecting placement revisions through iterative self-correction. $\bullet$ \textbf{Multimodal-Guided (Visual Feedback)} This variant furnishes the agent with rendered placement images as its sole inspection signal at each iteration, requiring the model to interpret spatial quality and identify layout deficiencies directly from visual observations rather than from structured numeric or symbolic outputs. $\bullet$ \textbf{Full-Feedback (All-Tools Adaptive)} This variant grants the agent unrestricted access to all available inspection tools \textemdash{} routability risk, overlap detection, out-of-boundary checking, and visual rendering \textemdash{} at each iteration, allowing the model to autonomously select the most informative feedback signals based on its current placement state and self-assessed quality, rather than being constrained to a single fixed feedback modality throughout the optimization loop.

\noindent\textbf{Diagnostic analysis protocol.}
Board complexity is measured by component count, while placement density is defined as the total area of pads, SMDs, and holes divided by the valid board area. For the component-count analysis, boards are grouped into six fixed intervals: 1\textendash{}20, 21\textendash{}40, 41\textendash{}60, 61\textendash{}100, 101\textendash{}160, and 161\textendash{}327 components. For the placement-density analysis, boards are grouped into 0\textendash{}10\%, 10\textendash{}20\%, 20\textendash{}30\%, 30\textendash{}40\%, 40\textendash{}60\%, and $\geq$60\% intervals. For each model and interval, we report the mean normalized overlap area over successfully parsed placements. Empty, invalid, or unparseable model responses are excluded; intervals containing no valid placements are reported as unavailable. For each model, the relative increase from the lowest to the highest interval is computed a
$100(\mathrm{OO}_{Q4}-\mathrm{OO}_{Q1})/\mathrm{OO}_{Q1}$. We report the median of these model-specific relative increases across models.

\begin{tcolorbox}[
    enhanced, breakable,
    colback=v1bg, colframe=v1border,
    boxrule=0.6pt, arc=0pt, outer arc=0pt,
    left=10pt, right=10pt, top=8pt, bottom=8pt,
]
\ttfamily\small
\#\textbf{Agentic\_prompt Variation 1 \textemdash{} HPWL-Guided (Routability Risk Feedback):}\\
You are an expert PCB layout engineer and placement-decision agent.\\
You must do two jobs in one response:\\
1. Produce the next candidate PCB placement.\\
2. Decide which repair actions and inspection tools should be used after
this candidate is saved and rendered.\\
\\
\textbf{Strict priority order:}\\
1. Return schema-valid JSON only. Include every required component exactly
once.\\
2. Repair deterministic geometry problems first: out\_of\_boundary, then
overlap.\\
3. Use HPWL/routability risk to guide compactness only after the placement
is geometrically valid.\\
\\
\textbf{Board Boundary:}\\
Extents: $x \in [x_0, x_1]$, $y \in [y_0, y_1]$ millimeters.\\
Exact outline primitives are in \texttt{board\_boundary.dxf\_details}
(summary: \{dxf\_summary\}).\\
\\
\textbf{Iteration:}\\
Current iteration: \{iteration\} / Maximum iterations: \{max\_iterations\}.\\
If iteration == max\_iterations, return your best candidate and do not
request unnecessary actions.\\
\\
\textbf{Required components} (exactly once each):\\
\{names\_json\}\\
\\
\textbf{Component Block Dimensions:}\\
The placement coordinate $(x, y)$ is the component CENTER in millimeters.\\
\texttt{block\_dimension} lists the physical blocks/pads/holes inside each
component.\\
Blocks from different components must not overlap on the same board side.\\
\quad - \textit{COMP\_NAME}: \textit{pad}:smd (front) $W$\,mm $\times$ $L$\,mm
at $(bx, by)$ rot=R0; $\ldots$\\
\\
\textbf{Source Rotation Operations} (read-only source orientation records):\\
\{source\_rotation\_operations\_json\}\\
\\
\textbf{Netlist:}\\
\quad - Net `\textit{NET\_NAME}': \textit{element}:\textit{pad}, $\ldots$\\
\\
\textbf{Safe Center Range Hints} (keep center inside these ranges per rotation):\\
\{fit\_hints\_json\}\\
\\
\textbf{Allowed rotation codes:}\\
\quad R0: front, 0\textdegree \quad R90: front, 90\textdegree \quad
R180: front, 180\textdegree \quad R270: front, 270\textdegree\\
\quad MR0: back mirrored, 0\textdegree \quad MR90: back mirrored, 90\textdegree
\quad MR180: back mirrored, 180\textdegree \quad MR270: back mirrored, 270\textdegree\\
\\
\textbf{Previous Context:}\\
\{previous\_context\_json\}\\
\\
\textbf{Recent History:}\\
\{history\_summary\_json\}\\
\\
\textbf{How to use previous context:}\\
- \texttt{previous\_inspection\_results} are observations from the previous
saved placement.\\
- If previous \texttt{hpwl} is high and geometry is clean, apply
\texttt{improve\_routability} to reduce net spread.\\
- \textbf{HPWL slot is replaced by \texttt{routability\_risk}}: a lower score
is better; compare current vs.\ previous score before requesting another
routability action.\\
- If geometry is still invalid, ignore routability risk until
out\_of\_boundary and overlap pass.\\
\\
\textbf{Available actions:}\\
\quad - \texttt{"fix\_boundary"}: repair blocks outside the board outline.\\
\quad - \texttt{"fix\_overlap"}: repair block overlaps between different
components.\\
\quad - \texttt{"improve\_routability"}: reduce routability risk / net spread
only after geometry is clean.\\
\quad - \texttt{"adjust\_visual\_layout"}: improve visual plausibility (not
applicable in this variation; omit or leave empty).\\
\quad - Return \texttt{action=[]} if no further iteration is needed.\\
\\
\textbf{Available inspection\_tools (exactly one allowed):}\\
\quad - \texttt{"hpwl"} \textemdash{} reports \texttt{routability\_risk} score for the
current candidate. Lower is better.\\
\\
\textbf{Placement rules:} (same as baseline \textemdash{} boundary, overlap, netlist
compactness)\\
\\
\textbf{Decision rules:}\\
1. If previous \texttt{hpwl} (routability\_risk) increased, action must
include \texttt{"improve\_routability"} with a concrete move plan.\\
2. \texttt{quality\_score} must not exceed 0.95 while known geometry
violations remain.\\
3. \texttt{visual\_feedback}: say \textit{``No previous visual image was
provided.''} (visualize is not available in this variation).\\
\\
\textbf{Output Format} (strictly follow this):\\
Return ONLY a valid JSON object, no markdown fences.\\
\{\\
\quad ``placements'': [\{\ldots\}],\\
\quad ``decision'': \{\\
\quad\quad ``action'': [\,],\\
\quad\quad ``inspection\_tools'': [``hpwl''],\\
\quad\quad ``quality\_score'': 1.0,\\
\quad\quad ``next\_action\_description'': ``\ldots'',\\
\quad\quad ``feedback'': ``\ldots'',\\
\quad\quad ``thought'': ``\ldots'',\\
\quad\quad ``visual\_feedback'': ``No previous visual image was provided.''\\
\quad \}\\
\}
\end{tcolorbox}

\vspace{1em}

\begin{tcolorbox}[
    enhanced, breakable,
    colback=v2bg, colframe=v2border,
    boxrule=0.6pt, arc=0pt, outer arc=0pt,
    left=10pt, right=10pt, top=8pt, bottom=8pt,
]
\ttfamily\small
\#\textbf{Agentic\_prompt Variation 2 \textemdash{} Overlap-Guided (Constraint Violation Feedback):}\\
You are an expert PCB layout engineer and placement-decision agent.\\
You must do two jobs in one response:\\
1. Produce the next candidate PCB placement.\\
2. Decide whether the current candidate needs a repair iteration based
exclusively on block overlap violations.\\
\\
\textbf{Strict priority order:}\\
1. Return schema-valid JSON only. Include every required component exactly
once.\\
2. Eliminate all block overlaps between different components.\\
3. Boundary compliance and routability are secondary; focus on
overlap-free placement above all else.\\
\\
\textbf{Board Boundary:}\\
Extents: $x \in [x_0, x_1]$, $y \in [y_0, y_1]$ millimeters.\\
Outline summary: \{dxf\_summary\}.\\
\\
\textbf{Iteration:}\\
Current iteration: \{iteration\} / Maximum iterations: \{max\_iterations\}.\\
\\
\textbf{Required components, Component Block Dimensions, Source Rotation\\
Operations, Netlist, Safe Center Range Hints, Allowed rotation codes:}\\
(identical to baseline \textemdash{} see shared section above)\\
\\
\textbf{Previous Context:}\\
\{previous\_context\_json\}\\
\\
\textbf{Recent History:}\\
\{history\_summary\_json\}\\
\\
\textbf{How to use previous context:}\\
- If previous \texttt{overlap} tool returned \texttt{ok: false}, the errors
list names the exact component pairs that collide; move those pairs apart
and include \texttt{"fix\_overlap"} in \texttt{decision.action}.\\
- Do not repeat a placement that failed overlap inspection unless you
have explicitly changed the relevant component positions.\\
\\
\textbf{Available actions:}\\
\quad - \texttt{"fix\_overlap"}: repair block overlaps between different components.\\
\quad - \texttt{"fix\_boundary"}: secondary; use only if overlap is already clean.\\
\quad - \texttt{"improve\_routability"}: secondary; use only if overlap is clean.\\
\quad - Return \texttt{action=[]} if \texttt{overlap} is expected to pass.\\
\\
\textbf{Available inspection\_tools (exactly one allowed):}\\
\quad - \texttt{"overlap"} \textemdash{} deterministic local block overlap detection.
Returns per-pair error messages when \texttt{ok: false}.\\
\\
\textbf{Decision rules:}\\
1. \texttt{quality\_score} must not exceed 0.95 while any overlap error
remains unresolved.\\
2. \texttt{visual\_feedback}: say \textit{``No previous visual image was
provided.''} (visualize is not available in this variation).\\
3. \texttt{feedback} must quote the specific component names from the
previous overlap error list and describe the move applied.\\
\\
\textbf{Output Format} (strictly follow this):\\
Return ONLY a valid JSON object, no markdown fences.\\
\{\\
\quad ``placements'': [\{\ldots\}],\\
\quad ``decision'': \{\\
\quad\quad ``action'': [\,],\\
\quad\quad ``inspection\_tools'': [``overlap''],\\
\quad\quad ``quality\_score'': 1.0,\\
\quad\quad ``next\_action\_description'': ``\ldots'',\\
\quad\quad ``feedback'': ``\ldots'',\\
\quad\quad ``thought'': ``\ldots'',\\
\quad\quad ``visual\_feedback'': ``No previous visual image was provided.''\\
\quad \}\\
\}
\end{tcolorbox}

\vspace{1em}

\begin{tcolorbox}[
    enhanced, breakable,
    colback=v3bg, colframe=v3border,
    boxrule=0.6pt, arc=0pt, outer arc=0pt,
    left=10pt, right=10pt, top=8pt, bottom=8pt,
]
\ttfamily\small
\#\textbf{Agentic\_prompt Variation 3 \textemdash{} Boundary-Guided (Out-of-Boundary Feedback):}\\
You are an expert PCB layout engineer and placement-decision agent.\\
You must do two jobs in one response:\\
1. Produce the next candidate PCB placement.\\
2. Decide whether the current candidate needs a repair iteration based
exclusively on out-of-boundary violations.\\
\\
\textbf{Strict priority order:}\\
1. Return schema-valid JSON only. Include every required component exactly
once.\\
2. Keep every transformed block strictly inside the board outline polygon.\\
3. Overlap avoidance and routability are secondary; spatial legality is
the single convergence criterion in this variation.\\
\\
\textbf{Board Boundary:}\\
Extents: $x \in [x_0, x_1]$, $y \in [y_0, y_1]$ millimeters.\\
Exact polygon outline is in \texttt{board\_boundary.dxf\_details}
(summary: \{dxf\_summary\}).\\
\textbf{CRITICAL}: use the polygon outline, not just the bounding box, for
boundary legality; a component center inside the bbox can still have blocks
outside an irregular outline.\\
\\
\textbf{Iteration:}\\
Current iteration: \{iteration\} / Maximum iterations: \{max\_iterations\}.\\
\\
\textbf{Required components, Component Block Dimensions, Source Rotation\\
Operations, Netlist, Safe Center Range Hints, Allowed rotation codes:}\\
(identical to baseline \textemdash{} see shared section above)\\
\\
\textbf{Previous Context:}\\
\{previous\_context\_json\}\\
\\
\textbf{Recent History:}\\
\{history\_summary\_json\}\\
\\
\textbf{How to use previous context:}\\
- If previous \texttt{out\_of\_boundary} tool returned \texttt{ok: false},
the errors list reports each violating block with its
\texttt{oob\_area} value; move the component center into the
\texttt{safe\_center\_ranges} for its rotation and include
\texttt{"fix\_boundary"} in \texttt{decision.action}.\\
- \texttt{SAFE\_CENTER\_RANGE\_HINTS} gives pre-computed safe center ranges
per rotation per component; prefer those ranges to avoid polygon violations.\\
\\
\textbf{Available actions:}\\
\quad - \texttt{"fix\_boundary"}: move violating component centers into their
safe range so all blocks lie inside the board polygon.\\
\quad - \texttt{"fix\_overlap"}: secondary; use only if boundary is already
clean.\\
\quad - \texttt{"improve\_routability"}: secondary; use only if boundary is
clean.\\
\quad - Return \texttt{action=[]} if \texttt{out\_of\_boundary} is expected
to pass.\\
\\
\textbf{Available inspection\_tools (exactly one allowed):}\\
\quad - \texttt{"out\_of\_boundary"} \textemdash{} deterministic polygon boundary
detection. Reports per-block \texttt{oob\_area} when \texttt{ok: false}.\\
\\
\textbf{Decision rules:}\\
1. \texttt{quality\_score} must not exceed 0.95 while any boundary error
remains unresolved.\\
2. \texttt{visual\_feedback}: say \textit{``No previous visual image was
provided.''} (visualize is not available in this variation).\\
3. \texttt{feedback} must state the violating component names and which
rotation + center coordinate was chosen to move them inside bounds.\\
\\
\textbf{Output Format} (strictly follow this):\\
Return ONLY a valid JSON object, no markdown fences.\\
\{\\
\quad ``placements'': [\{\ldots\}],\\
\quad ``decision'': \{\\
\quad\quad ``action'': [\,],\\
\quad\quad ``inspection\_tools'': [``out\_of\_boundary''],\\
\quad\quad ``quality\_score'': 1.0,\\
\quad\quad ``next\_action\_description'': ``\ldots'',\\
\quad\quad ``feedback'': ``\ldots'',\\
\quad\quad ``thought'': ``\ldots'',\\
\quad\quad ``visual\_feedback'': ``No previous visual image was provided.''\\
\quad \}\\
\}
\end{tcolorbox}

\vspace{1em}

\begin{tcolorbox}[
    enhanced, breakable,
    colback=v4bg, colframe=v4border,
    boxrule=0.6pt, arc=0pt, outer arc=0pt,
    left=10pt, right=10pt, top=8pt, bottom=8pt,
]
\ttfamily\small
\#\textbf{Agentic\_prompt Variation 4 \textemdash{} Multimodal-Guided (Visual Feedback):}\\
You are an expert PCB layout engineer and placement-decision agent.\\
You must do two jobs in one response:\\
1. Produce the next candidate PCB placement.\\
2. Decide whether the current candidate needs a visual improvement
iteration based exclusively on the rendered placement image.\\
\\
\textbf{Strict priority order:}\\
1. Return schema-valid JSON only. Include every required component exactly
once.\\
2. Use the attached rendered image from the previous iteration as the
primary feedback signal; analyze spatial density, component clustering,
and obvious visual irregularities.\\
3. Numeric metrics (overlap counts, boundary violations, HPWL) are not
available in this variation; all quality judgements must be derived from
visual observation.\\
\\
\textbf{Board Boundary:}\\
Extents: $x \in [x_0, x_1]$, $y \in [y_0, y_1]$ millimeters.\\
Outline summary: \{dxf\_summary\}.\\
\\
\textbf{Iteration:}\\
Current iteration: \{iteration\} / Maximum iterations: \{max\_iterations\}.\\
If a rendered image is attached, it belongs to iteration \{iteration$-$1\};
your visual analysis applies to that image and your new placement must
address any deficiencies you observe.\\
\\
\textbf{Required components, Component Block Dimensions, Source Rotation\\
Operations, Netlist, Safe Center Range Hints, Allowed rotation codes:}\\
(identical to baseline \textemdash{} see shared section above)\\
\\
\textbf{Previous Context (including attached rendered image):}\\
\{previous\_context\_json\}\\
\textit{Note}: \texttt{previous\_visualize\_note} \textemdash{} a rendered image from
the previous iteration is attached. Any visual problems you identify
belong to that previous image; write your analysis in
\texttt{decision.visual\_feedback} for the current iteration and update
the new placement accordingly.\\
\\
\textbf{Recent History:}\\
\{history\_summary\_json\}\\
\\
\textbf{How to use previous context:}\\
- Inspect the attached rendered image for crowding, isolated components,
long diagonal ratsnest lines, and unbalanced empty zones.\\
- Move components to address the visual deficiencies you identify; include
\texttt{"adjust\_visual\_layout"} in \texttt{decision.action}.\\
- Do not request \texttt{overlap} or \texttt{out\_of\_boundary} tools;
they are not available in this variation.\\
\\
\textbf{Available actions:}\\
\quad - \texttt{"adjust\_visual\_layout"}: improve visual plausibility using
the previous rendered image.\\
\quad - Return \texttt{action=[]} if the visual layout appears satisfactory.\\
\\
\textbf{Available inspection\_tools (exactly one allowed):}\\
\quad - \texttt{"visualize"} \textemdash{} render the current candidate as a PNG image.
The rendered image becomes the visual feedback for the next iteration.\\
\\
\textbf{Decision rules:}\\
1. \texttt{visual\_feedback} is mandatory and must describe specific layout
deficiencies observed in the previous image (e.g., component clustering,
large empty zones, suspected overlapping bodies).\\
2. If no image was provided (first iteration), write \textit{``No previous
visual image was provided.''} and propose an initial spread layout.\\
3. \texttt{quality\_score} reflects your visual confidence only; do not
assume geometry correctness.\\
\\
\textbf{Output Format} (strictly follow this):\\
Return ONLY a valid JSON object, no markdown fences.\\
\{\\
\quad ``placements'': [\{\ldots\}],\\
\quad ``decision'': \{\\
\quad\quad ``action'': [\,],\\
\quad\quad ``inspection\_tools'': [``visualize''],\\
\quad\quad ``quality\_score'': 0.9,\\
\quad\quad ``next\_action\_description'': ``\ldots'',\\
\quad\quad ``feedback'': ``\ldots'',\\
\quad\quad ``thought'': ``\ldots'',\\
\quad\quad ``visual\_feedback'': ``Observed crowding near board center \ldots''\\
\quad \}\\
\}
\end{tcolorbox}

\vspace{1em}

\begin{tcolorbox}[
    enhanced, breakable,
    colback=v5bg, colframe=v5border,
    boxrule=0.6pt, arc=0pt, outer arc=0pt,
    left=10pt, right=10pt, top=8pt, bottom=8pt,
]
\ttfamily\small
\#\textbf{Agentic\_prompt Variation 5 \textemdash{} Full-Feedback (All-Tools Adaptive):}\\
You are an expert PCB layout engineer and placement-decision agent.\\
You must do two jobs in one response:\\
1. Produce the next candidate PCB placement.\\
2. Autonomously select the most informative inspection tools and repair
actions based on your current assessment of the placement quality.\\
\\
\textbf{Strict priority order:}\\
1. Return schema-valid JSON only. Include every required component exactly
once.\\
2. Repair deterministic geometry problems first: out\_of\_boundary, then
overlap.\\
3. Use visual feedback to improve obvious layout issues after deterministic
geometry is addressed.\\
4. Use HPWL/netlist compactness only after the placement is geometrically
valid.\\
\\
\textbf{Board Boundary:}\\
Extents: $x \in [x_0, x_1]$, $y \in [y_0, y_1]$ millimeters.\\
Exact outline primitives are in \texttt{board\_boundary.dxf\_details}
(summary: \{dxf\_summary\}).\\
\\
\textbf{Iteration:}\\
Current iteration: \{iteration\} / Maximum iterations: \{max\_iterations\}.\\
If iteration == max\_iterations, return your best candidate and do not
request unnecessary actions.\\
\\
\textbf{Required components, Component Block Dimensions, Source Rotation\\
Operations, Netlist, Safe Center Range Hints, Allowed rotation codes:}\\
(identical to baseline \textemdash{} see shared section above)\\
\\
\textbf{Previous Context:}\\
\{previous\_context\_json\}\\
\textit{Note}: if \texttt{"visualize"} was selected in the previous
iteration, a rendered image is attached. Any visual problems you identify
belong to that previous image; write your analysis in
\texttt{decision.visual\_feedback} and update the new placement
accordingly.\\
\\
\textbf{Recent History:}\\
\{history\_summary\_json\}\\
\\
\textbf{How to use previous context:}\\
- If previous \texttt{overlap} failed, move the involved components apart
and include \texttt{"fix\_overlap"} in \texttt{decision.action}.\\
- If previous \texttt{out\_of\_boundary} failed, move the violating
component center into its safe range and include \texttt{"fix\_boundary"}.\\
- If previous \texttt{hpwl} is high but geometry is still invalid, ignore
hpwl until geometry is clean.\\
- If a rendered image is attached, inspect it now and summarize visual
deficiencies in \texttt{decision.visual\_feedback}.\\
\\
\textbf{Available actions:}\\
\quad - \texttt{"fix\_boundary"}: repair blocks outside the board outline.\\
\quad - \texttt{"fix\_overlap"}: repair block overlaps between different
components.\\
\quad - \texttt{"adjust\_visual\_layout"}: improve visual plausibility using
the previous rendered image.\\
\quad - \texttt{"improve\_routability"}: reduce HPWL / net spread only after
geometry is clean.\\
\quad - Return \texttt{action=[]} if no further iteration is needed.\\
\\
\textbf{Available inspection\_tools (select all that apply):}\\
\quad - \texttt{"visualize"}: render the current candidate image for visual
feedback in the next iteration.\\
\quad - \texttt{"hpwl"}: compute routability / compactness metric. Lower
is better; compare with previous iterations.\\
\quad - \texttt{"overlap"}: deterministic block overlap detection.\\
\quad - \texttt{"out\_of\_boundary"}: deterministic polygon boundary
detection.\\
\\
\textbf{Tool selection rules:}\\
1. Select the tools needed to verify the risk areas of your current
candidate.\\
2. If geometry is uncertain, always include \texttt{overlap} and
\texttt{out\_of\_boundary}.\\
3. Add \texttt{visualize} when you want rendered feedback for the next
iteration.\\
4. Add \texttt{hpwl} only after geometry is confirmed clean.\\
5. If no further iteration is needed, select the minimal set that would
confirm the candidate's final quality.\\
\\
\textbf{Decision rules:}\\
1. If previous deterministic inspection failed, \texttt{action} must
include the matching repair action.\\
2. \texttt{quality\_score} must not exceed 0.95 while known overlap or
boundary failures remain unresolved.\\
3. \texttt{visual\_feedback} must analyze the previous rendered image if
one was provided; otherwise say \textit{``No previous visual image was
provided.''}\\
\\
\textbf{Output Format} (strictly follow this):\\
Return ONLY a valid JSON object, no markdown fences.\\
\{\\
\quad ``placements'': [\{\ldots\}],\\
\quad ``decision'': \{\\
\quad\quad ``action'': [\,],\\
\quad\quad ``inspection\_tools'': [``overlap'', ``out\_of\_boundary''],\\
\quad\quad ``quality\_score'': 1.0,\\
\quad\quad ``next\_action\_description'': ``Finish with current placement if
selected inspections pass.'',\\
\quad\quad ``feedback'': ``Concrete repair instructions if another iteration
is needed. Empty if done.'',\\
\quad\quad ``thought'': ``Concise reason for selected actions and tools.'',\\
\quad\quad ``visual\_feedback'': ``Visual analysis of the previous rendered
image, or `No previous visual image was provided.'\,''\\
\quad \}\\
\}
\end{tcolorbox}




\subsubsection{Implementation of Routing Tool}

We evaluate the generated PCB placements by testing whether they can support a complete downstream routing workflow after automated post-processing. For each generated placement, we first instantiate the component locations in the Eagle board file and then export the design to the routing pipeline. The resulting board is converted into a DSN-compatible format and routed using an autorouter without manual adjustment. All routing experiments are conducted under the same design-rule settings to ensure a fair comparison across placements. Specifically, the routing setup uses a two-layer board with routing enabled on both the top and bottom layers. The via/pad rule is configured with a round shape and a radius of $16~\mathrm{mil}$. The spacing rule follows the same clearance matrix for all experiments: the default clearance is $8~\mathrm{mil}$, the boundary clearance is $10~\mathrm{mil}$, and the clearances among vias, SMD pads, pins, and copper areas are set between $8$ and $10~\mathrm{mil}$ according to the design-rule matrix.

We use routing success as the main functional metric, since a placement is only practically useful if the nets can be completed under the board constraints. Specifically, we report the number of successfully processed placements, the number of placements that complete routing, the routed net count, the total net count, and the routing completion ratio. We also track failures at each stage, including invalid component placement and routing failure. These metrics quantify not only whether the model can generate syntactically valid placements, but also whether those placements are geometrically and electrically feasible for downstream PCB routing.

\end{document}